\documentclass[letterpaper]{article} 
\usepackage{aaai2026}  
\usepackage{times}  
\usepackage{helvet}  
\usepackage{courier}  
\usepackage[hyphens]{url}  
\usepackage{graphicx} 
\usepackage{natbib}  
\usepackage{caption} 
\usepackage{algorithm}
\usepackage{algorithmic}
\usepackage{amssymb}
\usepackage{hhline}
\usepackage{newfloat}
\usepackage{listings}
\DeclareCaptionStyle{ruled}{labelfont=normalfont,labelsep=colon,strut=off} 
\floatstyle{ruled}
\newfloat{listing}{tb}{lst}{}
\floatname{listing}{Listing}
\usepackage{amsmath}
\usepackage{todonotes}
\usepackage[table]{xcolor}
\usepackage{multirow}
\usepackage{booktabs}
\usepackage[most]{tcolorbox}
\usepackage{cleveref}

\title{Bot Meets Shortcut: How Can LLMs Aid in Handling

Unknown Invariance OOD Scenarios?}

\author {
    Shiyan Zheng\textsuperscript{\rm 1, 2},
    Herun Wan\textsuperscript{\rm 1},
    Minnan Luo\textsuperscript{\rm 1, 2}\thanks{Corresponding author},
    Junhang Huang\textsuperscript{\rm 3}
}
\affiliations {
    \textsuperscript{\rm 1}School of Computer Science and Technology, Xi'an Jiaotong University, China\\
    \textsuperscript{\rm 2}State Key Laboratory of Communication Content Cognition, China\\
    \textsuperscript{\rm 3}Beijing Institute of Technology, China\\
    2223515385@stu.xjtu.edu.cn,  minnluo@xjtu.edu.cn
}

\usepackage{bibentry}

\begin{document}

\maketitle

\begin{abstract}\label{abs:abstract}

While existing social bot detectors perform well on benchmarks, their robustness across diverse real-world scenarios remains limited due to unclear ground truth and varied misleading cues. In particular, the impact of shortcut learning, where models rely on spurious correlations instead of capturing causal task-relevant features, has received limited attention. To address this gap, we conduct an in-depth study to assess how detectors are influenced by potential shortcuts based on textual features, which are most susceptible to manipulation by social bots. We design a series of shortcut scenarios by constructing spurious associations between user labels and superficial textual cues to evaluate model robustness. Results show that shifts in irrelevant feature distributions significantly degrade social bot detector performance, with an average relative accuracy drop of 32\% in the baseline models. To tackle this challenge, we propose mitigation strategies based on large language models, leveraging counterfactual data augmentation. These methods mitigate the problem from data and model perspectives across three levels, including data distribution at both the individual user text and overall dataset levels, as well as the model’s ability to extract causal information. Our strategies achieve an average relative performance improvement of 56\% under shortcut scenarios.

\end{abstract}

\begin{links}
    \link{Code}{https://github.com/worfsmile/BotsMeetShortcut}
\end{links}
\begin{links}
    \link{Paper}{https://doi.org/10.1609/aaai.v40i2.37137}
\end{links}


\section{Introduction}\label{sec:introduction}

Social bot detection has become a significant research topic because of the rapid development of social networks. By consensus, social bots are automated accounts controlled by computer programs that mimic human behavior on social platforms~\citep{ferrara2016rise, cresci2020decade}. These accounts perform actions such as posting content, commenting, liking, and sharing, actively participating in digital social interactions. Given their substantial influence on information dissemination and public opinion shaping, social bots have attracted increasing attention from both academia and society~\citep{gallotti2020assessing, elmas2022characterizing}. With the continued advancement of social bot detection research, many deep learning based methods have achieved increasingly strong performance on benchmark datasets~\citep{Feng_Tan_Li_Luo_2022, liu2023botmoe, yang2024sebot, li2025ets}.

\begin{figure}[t]
  \centering
  \includegraphics[width=\linewidth]{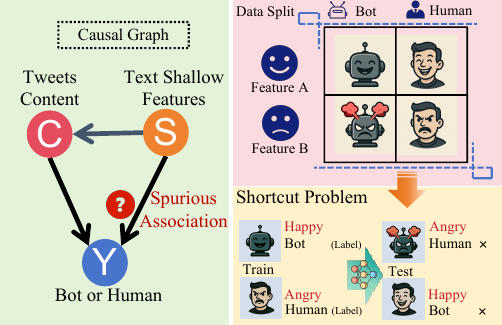}
  \caption{Schematic illustration of the shortcut scenario. As shown on the left, the causal graph depicts how spurious features (\emph{i.e.}, shortcuts) may interfere with inference, leading the model to learn incorrect reasoning from the training set. For instance, on the right, the users are partitioned after associating task-irrelevant feature  (\emph{e.g.}, sentiment) with their labels. As a result, the detector fails to generalize, and tends to make incorrect predictions when evaluated on diverse test instances.}
  \label{fig:fig1}
\end{figure}

However, social bot detection task has long been considered an ongoing arms race~\citep{davis2016botornot, cresci2017paradigm, cresci2020decade, feng2024does}. Social bots are not static adversaries but continuously evolve to evade detection, from early attempts to obscure user-level metadata~\citep{yang2020scalable}, to mimicking human-like language and retweeting genuine content~\citep{wei2019twitter, dukic2020you, feng2024does}, and more recently, to simulating complex social behaviors such as realistic follow networks and strategic user interactions~\citep{ali2019detect, feng2021botrgcn}. This evolutionary trajectory demonstrates the bots’ strong adaptability and adversarial nature, posing continuous challenges to static detection methods and underscoring the need for more robust and generalizable solutions~\citep{cresci2023demystifying}.

What’s more, despite the impressive performance of state-of-the-art models, numerous studies suggest that deep learning models often exploit spurious correlations or shortcuts, which are shallow and irrelevant features correlated with task labels but lacking causal relevance (\cref{fig:fig1}). This hinders their ability to learn truly meaningful representations~\citep{geirhos2020shortcut,wan2025truth}, raising concerns about their robustness and generalization. This challenge is particularly pronounced in the social bot detection task, as current detectors still struggle to generalize across different benchmarks, data distributions, and time due to biases in the datasets~\citep{5995347,hays2023simplistic}. Many models are closely tied to the specific datasets or network structures used during training~\citep{li2025ets}, limiting their effectiveness in dynamic, real-world scenarios. This challenge is largely driven by dataset biases and the ever-evolving nature of social network structures, discourse topics, and user behavior~\citep{cresci2020decade,cresci2023demystifying}, prompting a growing body of research to focus on this issue~\citep{mannocci2024detection, tardelli2024multifaceted, tardelli2024temporal}.

In this work, we conduct an in-depth investigation into the generalization ability of social bot detection models to evaluate how detectors are influenced by inherent shortcuts. Focusing on the intrinsic textual features of social media users, we design a series of distribution shift scenarios to examine whether shallow \textbf{text-level} perturbations can trigger shortcut learning in existing detection systems~\citep{geirhos2020shortcut, wan2025truth}. Building on these insights, we further propose a set of debiasing strategies both at the data and model levels, leveraging large language models (LLMs) to enhance robustness under unknown invariance conditions. Our main contributions are twofold:
\begin{itemize}
\item\textbf{Potential Shortcut in Social Bot Detection.}
We first investigate the shortcut learning problem in social bot detection by focusing on endogenous textual features, namely sentiment, topic, emotion, and human values. Inspired by the concept of spurious correlations in distribution shifts and shortcut learning~\citep{MORENOTORRES2012521, 8496795, geirhos2020shortcut}, we align labels with these superficial attributes to construct pseudo-correlated, biased scenarios. We partition the data based on textual feature to create train set and test set across three most authoritative and widely-used social bot detection datasets, Cresci-2015-Data~\citep{cresci2015fame}, Cresci-2017-Data~\citep{cresci2017paradigm}, and Twibot-20~\citep{feng2021twibot}. As the most commonly used detecting methods can be broadly categorized into text-based~\citep{luo2020deepbot, cai2024lmbot, feng2024does} and graph-based~\citep{ali2019detect, feng2021botrgcn, he2024botdgt} approaches, we evaluate the performance of representative baseline models under the given conditions. Our experimental results show a relative performance drop averaging $\textbf{33\%}$ for text-based models and $\textbf{30\%}$ for graph-based models, while commonly used debiasing methods remain largely ineffective in mitigating the degradation caused by such biases.

\item\textbf{Mitigation Methods.}
To address this challenge, we further explore counterfactual intervention approaches using LLMs~\citep{liu2021counterfactual,mishra2024llm} to generate \textbf{rewritten} text. Our proposed methods target bias reduction from three perspectives: the surface tendency of user text features, the construction of augmented datasets, and the representation ability of the language feature extraction model. Through these insights, we propose targeted and effective mitigation strategies to enhance robustness in the presence of spurious correlations, achieving an average relative performance improvement of $\textbf{59\%}$ on text-based models and $\textbf{53\%}$ on graph-based models compared with the shortcut setting before augmentation in our observations.
\end{itemize}

\begin{figure}[t]
\hspace{0cm}
\includegraphics[width=1\linewidth]{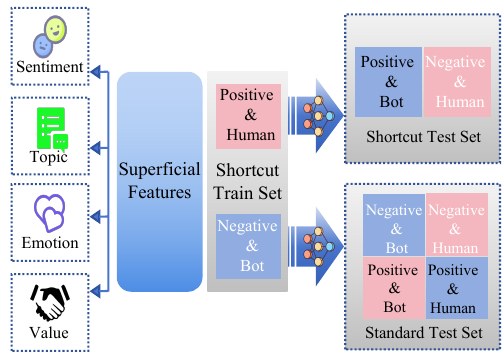}
  \caption{
A diagram of our shortcut settings.
We focus on the superficial features of the text, such as sentiment, topic, emotion, and human values, and set shortcuts to these features in the training set. In the test set, we either reverse the pseudo-correlation between features and the label in the shortcut test set or eliminate these shortcuts in the standard test set.
}
  \label{fig:fig2}
\end{figure}

\begin{table*}[t]
\centering
\begin{tabular}{ll|cc|cc|cc}
\toprule
\multicolumn{2}{c}{\multirow{2}{*}{RoBERTa}} & \multicolumn{2}{c}{{\small Cresci-2015-Data}} & \multicolumn{2}{c}{{\small Cresci-2017-Data}} & \multicolumn{2}{c}{{\small Twibot-20}} \\
 &  & {$\text{\small Shortcut}_{\text{te}}$} & {$\text{\small Standard}_{\text{te}}$} & {$\text{\small Shortcut}_{\text{te}}$} & {$\text{\small Standard}_{\text{te}}$} & {$\text{\small Shortcut}_{\text{te}}$} & {$\text{\small Standard}_{\text{te}}$} \\
\cmidrule(lr){1-8}
\multirow{3}{*}{Sentiments} & $\text{Standard}_{\text{tr}}$ & .945 & .955 & .884 & .900 & .682 & .685 \\
 & $\text{Shortcut}_{\text{tr}}$ & .565 & .780 & .287 & .625 & .051 & .523 \\
 & Difference & 40\%$\downarrow$ & 18\%$\downarrow$ & 67\%$\downarrow$ & 30\%$\downarrow$ & 92\%$\downarrow$ & 23\%$\downarrow$ \\
\cmidrule(lr){1-8}
\multirow{3}{*}{Emotions} & $\text{Standard}_{\text{tr}}$ & .982 & .979 & .963 & .940 & .684 & .687 \\
 & $\text{Shortcut}_{\text{tr}}$ & .849 & .917 & .555 & .766 & .110 & .520 \\
 & Difference & 13\%$\downarrow$ & 6\%$\downarrow$ & 42\%$\downarrow$ & 18\%$\downarrow$ & 83\%$\downarrow$ & 24\%$\downarrow$ \\
\cmidrule(lr){1-8}
\multirow{3}{*}{Topics} & $\text{Standard}_{\text{tr}}$ & .985 & .973 & .894 & .885 & .663 & .684 \\
 & $\text{Shortcut}_{\text{tr}}$ & .941 & .962 & .587 & .757 & .164 & .535 \\
 & Difference & 4\%$\downarrow$ & 1\%$\downarrow$ & 34\%$\downarrow$ & 14\%$\downarrow$ & 75\%$\downarrow$ & 21\%$\downarrow$ \\
\cmidrule(lr){1-8}
\multirow{3}{*}{Values} & $\text{Standard}_{\text{tr}}$ & .915 & .928 & .890 & .894 & .691 & .679 \\
 & $\text{Shortcut}_{\text{tr}}$ & .659 & .827 & .514 & .755 & .180 & .550 \\
 & Difference & 28\%$\downarrow$ & 10\%$\downarrow$ & 42\%$\downarrow$ & 15\%$\downarrow$ & 73\%$\downarrow$ & 18\%$\downarrow$ \\
\bottomrule
\end{tabular}
\caption{Accuracy of RoBERTa under standard and shortcut training and testing conditions. The vertical dimension denotes training configurations, $\text{Standard}_{\text{tr}}$ (randomly sampled, balanced) versus $\text{Shortcut}_{\text{tr}}$ (spuriously correlated shallow features), while the horizontal dimension denotes test feature distributions, $\text{Standard}_{\text{te}}$ (randomly sampled, balanced) versus $\text{Shortcut}_{\text{te}}$ (distributional shift). The difference between paired entries reported as the relative drop from standard to shortcut settings reflects the model’s vulnerability to spurious correlations.}
\label{tab:table1}
\end{table*}

\section{Potential Shortcuts}\label{sec:2potential-shortcuts}

Shortcut learning occurs when models exploit superficial features (\emph{e.g.}, sentiment cues in the social bot detection), which are often shallow and easier to capture~\citep{geirhos2020shortcut}. In the social bot detection task, the classification criteria are quite complex and therefore easily influenced by unrelated factors. We explore potential shortcut learning factors in \textbf{user text}, which is the most information-rich and bot-manipulable medium in social networks, and empirically demonstrate that social bot detection can be largely affected by shortcut learning through constructing endogenous shortcut scenarios (\cref{fig:fig2}).

\subsection{Shortcut Learning Scenarios Setup}
\label{sec:2-1shortcut-learning-scenario-setup}

Formulate the social bot detection as a binary classification task in which each instance corresponds to a user $u \in \mathcal{U}$ where $\mathcal{U}$ denotes the set of all users connected through the social network graph structure. Users are represented by the textual feature vector extracted from their selected posts and each instance has an associated label $y \in \{0,1\}$, indicating whether the user is a human $(y=0)$ or a bot $(y=1)$. To analyze user's textual features from different perspectives, we denote $\phi^{\text{causal}}_{\text{task}} (u)$ as the semantic signals that are causally related to the user label $y$, while $\phi^{\text{spu}}_{\text{fea}} (u)$ indicates the textual attribute associated with certain spurious factors (\emph{e.g.}, text emotion, topic and so on) that may correlate with $y$ while lacking causal relevance.

Let $\mathcal{S}_{\text{fea}}$ denote the set of possible values of $\phi_{\text{fea}}^{\text{spu}}(u)$ associated with some specific feature (\emph{e.g.}, emotion) and partition this set into two mutually exclusive and internally similar subsets: $\mathcal{S}^{\text{pos}}_{\text{fea}}$ and $\mathcal{S}^{\text{neg}}_{\text{fea}}$. Based on this partition, we define two corresponding instance sets as
\begin{align*}
\mathcal{U}^{\text{pos}}_{\text{fea}} &= \{u\ \mid\  \phi^{\text{spu}}_{\text{fea}}(u)\  \in \ \mathcal{S}^{\text{pos}}_{\text{fea}} \},\\
\mathcal{U}^{\text{neg}}_{\text{fea}} &= \{u\ \mid \ \phi^{\text{spu}}_{\text{fea}}(u)\  \in \ \mathcal{S}^{\text{neg}}_{\text{fea}} \}.
\end{align*}

We collect the instances into set $\mathcal{D} = \{ (u_i, y_i)\}_{i=1}^n$ and split it as the \textbf{Shortcut Train Set} $\mathcal{D}^{\text{str}}_{\text{fea}}$ and \textbf{Shortcut Test Set} $\mathcal{D}^{\text{ste}}_{\text{fea}}$ based on the presence of special \emph{spurious feature}, \emph{i.e.},
\begin{align*}
\mathcal{D}^{\text{str}}_{\text{fea}} =&\{ (u_{i}, y_{i}) \in D \mid u_{i} \in \ \mathcal{U}^{\text{pos}}_{\text{fea}},\ y_{i} = 1 \} \cup\\
&\{ (u_{i}, y_{i}) \in D \mid u_{i} \in \ \mathcal{U}^{\text{neg}}_{\text{fea}},\ y_{i} = 0 \},\\
\mathcal{D}^{\text{ste}}_{\text{fea}} = &\{ (u_{i}, y_{i}) \in D \mid u_{i} \in \ \mathcal{U}^{\text{pos}}_{\text{fea}},\ y_{i} = 0 \} \cup\\
&\{ (u_{i}, y_{i}) \in D \mid u_{i} \in \ \mathcal{U}^{\text{neg}}_{\text{fea}},\ y_{i} = 1 \}.
\end{align*}

For example, one may train on happy bots (sampled from $\mathcal{U}^{\text{pos}}_{\text{emotion}}$ labeled 1) and angry humans (sampled from $\mathcal{U}^{\text{neg}}_{\text{emotion}}$ labeled 0), then test on angry bots and happy humans in the \textbf{Shortcut Test Set}. Note that in the \textbf{Standard Test Set}, labels are \emph{independent} of shortcut features.

In this work, we specifically focus on several types of shallow textual features that could potentially trigger shortcut learning: sentiment, topic, emotion, and human values.
For instance, in the case of sentiment, we assign users whose texts express a clearly \textit{positive} tone to $\mathcal{U}^{\text{pos}}_{\text{senti}}$, and those with a \textit{negative} tone to $\mathcal{U}^{\text{neg}}_{\text{senti}}$. For topic, those texts discussing \textit{daily life} are assigned to $\mathcal{U}^{\text{pos}}_{\text{topic}}$, while those pertaining to \textit{pop culture} and \textit{sports} fall into $\mathcal{U}^{\text{neg}}_{\text{topic}}$. Specifically, we assign users whose texts corresponding feature types cannot be clearly categorized into the above-defined sets to $\mathcal{U}^{\text{neu}}_{\text{fea}}$. Based on this subset division, we obtain train set and test set accordingly to construct our shortcut learning setting through text filtering (more details are provided in \cref{app:C-detailed-scenario-design}).

\begin{figure*}[t]
\hspace{0cm}
\includegraphics[width=1\linewidth]{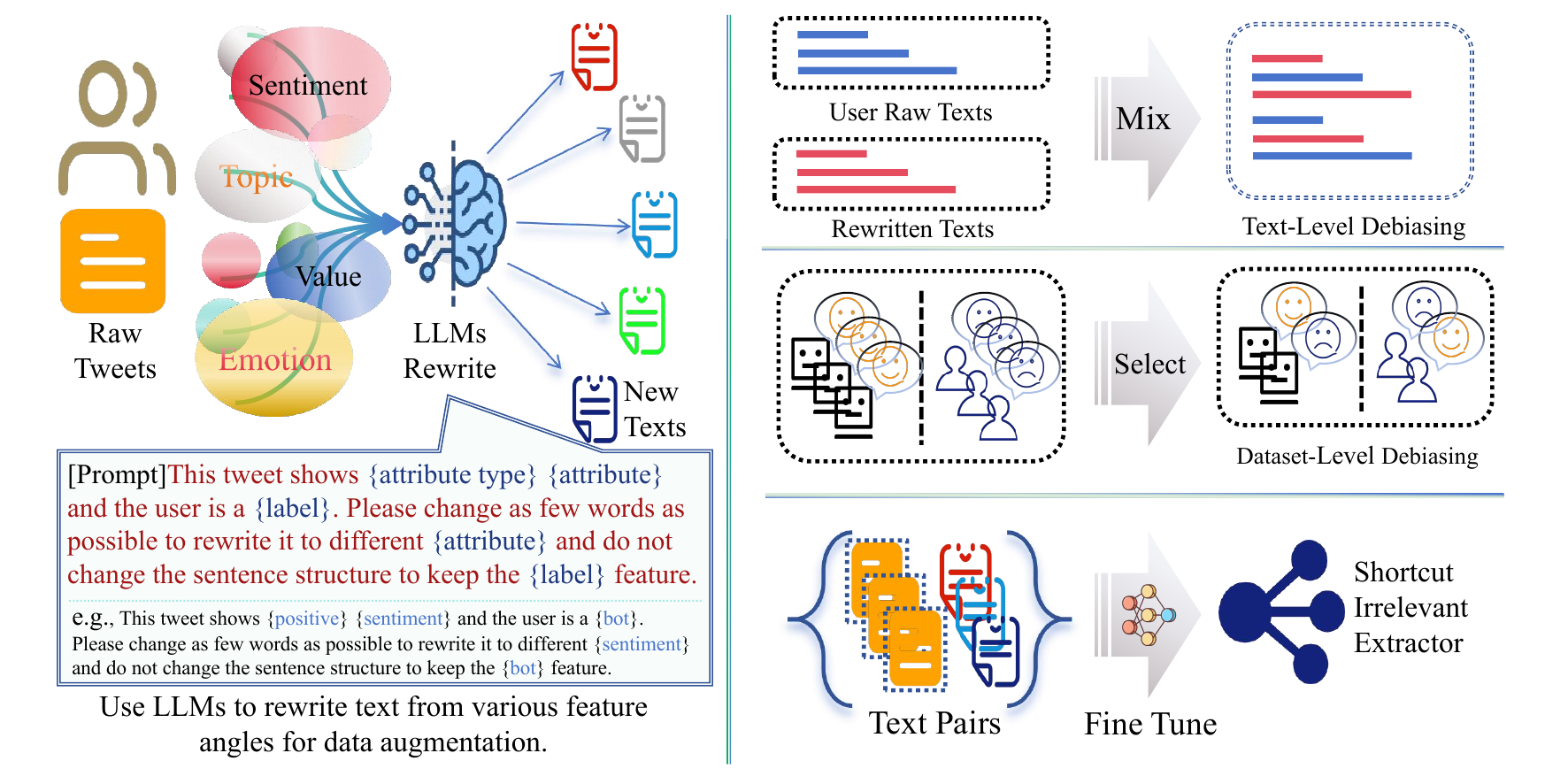}
\caption{Overview of our shortcut learning mitigation framework. The left side illustrates how LLMs are used to augment data by rewriting text from different attribute perspectives, while preserving the user’s label-related semantics. The right part demonstrates the mitigation process at three levels: balancing the semantic content of individual users’ texts at the user level, balancing feature distributions across classes at the dataset level, and enhancing the feature extractor's ability to capture causal information across different shortcut shifts at the language model embedding level by employing contrastive learning.}
\label{fig:fig3}
\end{figure*}

\subsection{Impact on Classifiers}\label{sec:2-2impact-on-classifiers}

We test the impact of the above-mentioned scenarios on the detectors. We use some of the most well-known and widely used social bot detection datasets, Cresci-2015-Data~\citep{cresci2015fame}, Cresci-2017-Data~\citep{cresci2017paradigm}, and Twibot-20~\citep{feng2021twibot}. For these datasets, we design shortcut learning scenarios and evaluate three different approaches: 
\begin{itemize}
\item \textbf{Based on Language Models (LM):} We use the RoBERTa~\citep{liu2019roberta} model with frozen parameters to generate feature embeddings for the user's text. We then use an MLP classifier to classify the text.
\item \textbf{Based on Graph:} We use the embedded text as features and incorporate social network graph structure information, applying BotRGCN~\citep{feng2021botrgcn} classifier for classification.
\item \textbf{Based on Debiasing Approaches:} We try to employ debiasing methods to reduce the effect of shortcut features on classification. We tested a representative causal decoupling-based debiasing model, CIGA~\citep{chen2022learning}, by representing tweets with abstract meaning representations (AMR)~\citep{banarescu2013abstract}.
\end{itemize}

The experimental results, as shown in \cref{tab:table1,app:D-potential-shortcut}, indicate that all of these methods were affected by the shortcut learning scenario. For example, in the text-based method, compared to the normal distribution scenario (\emph{i.e.}, training on the standard train set), the relative classification accuracy in \textit{shortcut test setting} on average decreases by $\textbf{50\%}$, with the largest decrease being $\textbf{92\%}$. In \textit{standard test setting}, the average relative accuracy drop is $\textbf{17\%}$, with the most significant decrease being $\textbf{30\%}$, which demonstrates that existing model architectures suffer significant performance degradation. And the debiasing model (we implement it by combining AMR’s~\citep{banarescu2013abstract} explicit semantic graphs with CIGA’s~\cite{chen2022learning} causal disentanglement) is not effective enough to alleviate this issue, as it exhibits poor performance in the shortcut scenarios (compared in \cref{tab:table2} and detail in \cref{app:D-potential-shortcut}).

\section{Mitigation Methods}\label{sec:3mitigation-methods}

Leveraging the competency of LLMs in social bot detection tasks~\citep{feng2024does}, we employ a counterfactual data augmentation (CDA) strategy comprising two main components: (1) At the first step, we perform CDA on the \textbf{Training Set} to generate texts that reverse specific biased feature while preserving the main semantic content and label criteria. (2) We mitigate shortcut-inducing correlations at three levels, including semantic patterns in the \textit{text level}, skewed feature distributions in the \textit{dataset level}, and embedding language model's debiasing ability in the \textit{model level} (\cref{fig:fig3}).

\subsection{Counterfactual Data Augmentation}\label{sec:3-1counterfactual-data-augmentation}

Specialized LLMs can identify potential superficial feature biases \textbf{in training data} by analyzing text patterns (\emph{e.g.}, clearly positive tone). Then to counteract these spurious correlations, we employ a prompt-based LLM \textbf{rewriting} method (using DeepSeek API~\citep{guo2025deepseek} in our implementation) to generate counterfactual text while maintaining semantic consistency. Specifically, we prompt the model to alter the expression of specified shallow features (\emph{e.g.}, sentiment, human values and topic cues) \emph{without} changing the core meaning and sentence structure of the original texts, while preserving the distinguishing characteristics between social bots and humans in the original texts (some examples in \cref{app:E-LLMs-modify-example}). And by rewriting an original text $T_{\text{raw}}$ into a new version $T_{\text{new}}$, we obtain a text pair $(T_{\text{raw}}, T_{\text{new}})$.

\begin{table*}[t]
\centering
\begin{tabular}{ll|cc|cc|cc}
\toprule
\multicolumn{2}{c}{\multirow{2}{*}{RoBERTa}} & \multicolumn{2}{c}{{\small Cresci-2015-Data}} & \multicolumn{2}{c}{{\small Cresci-2017-Data}} & \multicolumn{2}{c}{{\small Twibot-20}} \\
 &  & {\small $\text{Shortcut}_{\text{te}}$} & {\small $\text{Standard}_{\text{te}}$} & {\small $\text{Shortcut}_{\text{te}}$} & {\small $\text{Standard}_{\text{te}}$} & {\small $\text{Shortcut}_{\text{te}}$} & {\small $\text{Standard}_{\text{te}}$} \\
\cmidrule(lr){1-8}
\multirow{5}{*}{Sentiments} & $\text{Shortcut}_{\text{tr}}$ & .565 & .780 & .287 & .625 & .051 & .523 \\
 & AMR+CIGA & .678\;(19\%$\uparrow$) & .838\;(7\%$\uparrow$) & .271\;(5\%$\downarrow$) & .671\;(7\%$\uparrow$) & .112\;(120\%$\uparrow$) & .531\;(1\%$\uparrow$) \\
 & $\text{Text-Level}^{*}$ & .840\;(48\%$\uparrow$) & .910\;(16\%$\uparrow$) & .597\;(108\%$\uparrow$) & .757\;(21\%$\uparrow$) & .229\;(349\%$\uparrow$) & .559\;(6\%$\uparrow$) \\
 & $\text{Dataset-Level}^{*}$ & .835\;(47\%$\uparrow$) & \underline{.912\;(16\%$\uparrow$)} & .667\;(132\%$\uparrow$) & .794\;(27\%$\uparrow$) & .288\;(466\%$\uparrow$) & .571\;(9\%$\uparrow$) \\
 & $\text{Model-Level}^{*}$ & \underline{.864\;(52\%$\uparrow$)} & .907\;(16\%$\uparrow$) & \underline{.690\;(140\%$\uparrow$)} & \underline{.836\;(33\%$\uparrow$)} & \underline{.415\;(715\%$\uparrow$)} & \underline{.580\;(10\%$\uparrow$)} \\
\cmidrule(lr){1-8}
\multirow{5}{*}{Emotions} & $\text{Shortcut}_{\text{tr}}$ & .849 & .917 & .555 & .766 & .110 & .520 \\
 & AMR+CIGA & .770\;(9\%$\downarrow$) & .870\;(5\%$\downarrow$) & .382\;(31\%$\downarrow$) & .679\;(11\%$\downarrow$) & .232\;(109\%$\uparrow$) & .538\;(3\%$\uparrow$) \\
 & $\text{Text-Level}^{*}$ & \underline{.936\;(10\%$\uparrow$)} & .964\;(5\%$\uparrow$) & \underline{.670\;(20\%$\uparrow$)} & \underline{.819\;(6\%$\uparrow$)} & .235\;(113\%$\uparrow$) & .569\;(9\%$\uparrow$) \\
 & $\text{Dataset-Level}^{*}$ & \underline{.936\;(10\%$\uparrow$)} & \underline{.971\;(5\%$\uparrow$)} & .610\;(9\%$\uparrow$) & .782\;(2\%$\uparrow$) & .288\;(160\%$\uparrow$) & .569\;(9\%$\uparrow$) \\
 & $\text{Model-Level}^{*}$ & .923\;(8\%$\uparrow$) & .933\;(1\%$\uparrow$) & .628\;(13\%$\uparrow$) & .803\;(4\%$\uparrow$) & \underline{.400\;(262\%$\uparrow$)} & \underline{.585\;(12\%$\uparrow$)} \\
\cmidrule(lr){1-8}
\multirow{5}{*}{Topics} & $\text{Shortcut}_{\text{tr}}$ & .941 & .962 & .587 & .757 & .164 & .535 \\
 & AMR+CIGA & .774\;(17\%$\downarrow$) & .850\;(11\%$\downarrow$) & .405\;(30\%$\downarrow$) & .656\;(13\%$\downarrow$) & .189\;(15\%$\uparrow$) & .530\;(0\%$\downarrow$) \\
 & $\text{Text-Level}^{*}$ & .979\;(4\%$\uparrow$) & \underline{.972\;(1\%$\uparrow$)} & .780\;(32\%$\uparrow$) & .833\;(9\%$\uparrow$) & .296\;(80\%$\uparrow$) & .577\;(7\%$\uparrow$) \\
 & $\text{Dataset-Level}^{*}$ & \underline{.992\;(5\%$\uparrow$)} & .962\;(0\%$\downarrow$) & .807\;(37\%$\uparrow$) & .807\;(6\%$\uparrow$) & \underline{.607\;(270\%$\uparrow$)} & .567\;(5\%$\uparrow$) \\
 & $\text{Model-Level}^{*}$ & .954\;(1\%$\uparrow$) & .953\;(0\%$\downarrow$) & \underline{.839\;(42\%$\uparrow$)} & \underline{.867\;(14\%$\uparrow$)} & .522\;(218\%$\uparrow$) & \underline{.635\;(18\%$\uparrow$)} \\
\cmidrule(lr){1-8}
\multirow{5}{*}{Values} & $\text{Shortcut}_{\text{tr}}$ & .659 & .827 & .514 & .755 & .180 & .550 \\
 & AMR+CIGA & .592\;(10\%$\downarrow$) & .785\;(5\%$\downarrow$) & .572\;(11\%$\uparrow$) & .739\;(2\%$\downarrow$) & .225\;(24\%$\uparrow$) & .536\;(2\%$\downarrow$) \\
 & $\text{Text-Level}^{*}$ & \underline{.762\;(15\%$\uparrow$)} & \underline{.873\;(5\%$\uparrow$)} & .711\;(38\%$\uparrow$) & .826\;(9\%$\uparrow$) & .295\;(63\%$\uparrow$) & .567\;(3\%$\uparrow$) \\
 & $\text{Dataset-Level}^{*}$ & .633\;(3\%$\downarrow$) & .818\;(1\%$\downarrow$) & .771\;(50\%$\uparrow$) & .828\;(9\%$\uparrow$) & .407\;(126\%$\uparrow$) & .595\;(8\%$\uparrow$) \\
 & $\text{Model-Level}^{*}$ & .615\;(6\%$\downarrow$) & .812\;(1\%$\downarrow$) & \underline{.830\;(61\%$\uparrow$)} & \underline{.869\;(15\%$\uparrow$)} & \underline{.488\;(171\%$\uparrow$)} & \underline{.614\;(11\%$\uparrow$)} \\
\bottomrule
\end{tabular}
\caption{Mitigation effects of our methods on RoBERTa model ( * indicates our strategies). We compare our mitigation strategies against the original baseline and representative debiasing methods under the shortcut setting, and report the relative improvement over the original shortcut scenario. The best performance in each group is highlighted with an underline. Results demonstrate that our methods effectively alleviate the impact of shortcuts, and in most cases, the performance approaches or even matches that of the standard setting.}
\label{tab:table2}
\end{table*}

\subsection{Text-Level Debiasing}\label{sec:3-2text-level-debiasing}

Each user $u$ is associated with a set of tweets $\mathcal{T} = \{T_1, T_2, \dots, T_k\}$ ($k=5$ in our experiments), and suppose these texts exhibit a specific bias under a shallow feature attribute. We prompt the LLM to rewrite each tweet such that the resulting set $\mathcal{T'} = \{T_1', T_2', \dots, T_k'\}$ presents a different or neutral tendency with respect to the attribute, while preserving the original label inference basis.
We then combine original and rewritten texts to select a balanced subset with minimized bias. To this end, we define a feature bias score as
\begin{equation}
f = \left| \frac{e^{R_{\text{pos}}} - e^{R_{\text{neg}}}}{e^{R_{\text{neu}}}} \right|,
\label{eq:balance_score}
\end{equation}
where $R_{\text{pos}}$, $R_{\text{neg}}$, and $R_{\text{neu}}$ represent proportions of tweets with positive, negative expressions of the attribute, and $R_\text{neu}$ is proportion of types out of them (\emph{i.e.,} defined in \cref{sec:2-1shortcut-learning-scenario-setup}). 
We evaluate all combinations of original/revised texts (using binary selection vector of size $k$), and select the one minimizing $f$ to obtain a debiased tweet set $\mathcal{T''} = \{T_1'', T_2'', \dots, T_k''\}$ for each user (\cref{alg:algorithm1} in \cref{app:F-algorithms}).

\subsection{Dataset-Level Debiasing}\label{sec:3-3dataset-level-debiasing}

After obtaining the bias-mitigated texts at the user level, we also conduct dataset-level debiasing to reduce the shortcut learning between labels and superficial features.

For each label class $y \in \{0, 1\}$, we randomly divide its corresponding samples in the training set: half use the original tweets $\mathcal{T}$, and the other half use the rewritten tweets $\mathcal{T'}$. Due to the limitations of LLMs in precisely controlling rewriting directions and the inherent directional bias in the original text, not all samples can be reliably modified as expected, meaning that not all users can obtain $\mathcal{T'}$. In cases where the samples cannot be evenly split, the remaining instance is assigned the mixed version $\mathcal{T''}$. This strategy ensures a relatively balanced distribution of shallow features across different classes and weakens the model's reliance on spurious correlations (\cref{alg:algorithm2} in \cref{app:F-algorithms}).

\subsection{Model-Level Mitigation}\label{sec:3-4model-level-mitigation}

On top of getting counterfactual texts, we further fine-tune the language feature extractor using contrastive learning to enhance its ability to capture causal information.

\paragraph{Data Preparing}
From the CDA steps, we obtain a collection of text pairs $\{ (T_{\text{raw}}, T_{\text{new}})\}$, where each pair consists of an original text and its rewritten version. And we collected over 200,000 such pairs, which serve as the dataset for our fine-tuning process.

\paragraph{Debiasing via Contrastive Fine-Tuning}  
To remove shortcut features from text embeddings, we fine-tune a pretrained language feature extractor model $M$ (we use RoBERTa~\citep{liu2019roberta} in our experiment) using a joint objective that encourages manifold preservation and suppresses spurious information. Denote $M_{\text{raw}}$ as the model before finetuning, and $M_\text{{finetune}}$ as the model after finetuning. Given a text pair $(T_{\text{raw}}, T_{\text{new}})$, we compute the embedding of the raw text as $h_{\text{raw}} = M_{\text{raw}} (T_{\text{raw}})$, $h_{\text{pos}} = M_\text{{finetune}} (T_{\text{raw}})$, and, where applicable, a contrastive example $h_{\text{neg}} = M_\text{{finetune}} (T_{\text{new}})$. The overall loss to be minimized is defined as
\begin{equation}
\mathcal{L} = \mathcal{L}_{\text{manifold}} + \lambda \mathcal{L}_{\text{MI}}.
\end{equation}
where $\lambda$ is a hyperparameter that balances the contribution of the mutual information loss relative to the manifold preservation loss.

The manifold loss preserves semantic structure across texts by aligning the similarity distributions between raw and transformed embeddings, that is a modified version of that introduced by~\citet{park2019relational}, \emph{i.e.},
\begin{align}
\mathcal{L}_{\text{manifold}} &= \lambda_1 \mathcal{L}_{\text{positive}} + \lambda_2 \mathcal{L}_{\text{negative}},\\
\mathcal{L}_{\text{positive}} &= KL (\text{Sim} (H_{\text{pos}}), \text{Sim} (H_{\text{raw}})),\\
\mathcal{L}_{\text{negative}} &= KL (\text{Sim} (H_{\text{neg}}), \text{Sim} (H_{\text{raw}})),
\end{align}
where $\lambda_{1},\lambda_2$ denotes a hyperparameter. $KL (\cdot,\cdot)$ denotes the Kullback-Leibler (KL) divergence. And $\text{Sim}(H)$ denotes cosine similarities computed between $h$ representations within a batch.

To remove shortcut signals, we maximize the mutual information (MI) between positive and negative examples, \emph{i.e.},
\begin{equation}
\mathcal{L}_{\text{MI}} = - I (h_{\text{pos}}, h_{\text{neg}}).
\end{equation}
We maximize mutual information by maximizing its lower bound, utilizing the InfoNCE method, shown in \cref{eq:InfoNCE} proposed by~\citet{oord2018representation}.
\begin{align}
\label{eq:InfoNCE}
&I_{\text{NCE}} = \frac{1}{N} \sum_{i=1}^{N} \log \frac{e^{f (h_{\text{pos}}^{i}, h_{\text{neg}}^{j})}}{\frac{1}{N} \sum_{j=1}^{N} e^{f (h_{\text{pos}}^{i}, h_{\text{neg}}^{j})}}  \\
= &\frac{1}{N} \sum_{i=1}^{N} f (h_{\text{pos}}^{i}, h_{\text{neg}}^{i}) - \frac{1}{N} \sum_{i=1}^{N} \left[ \log \frac{1}{N} \sum_{j=1}^{N} e^{f(h_{\text{pos}}^{i}, h_{\text{neg}}^{j})} \right].\nonumber
\end{align}
Here, $f$ is a learnable non-negative scoring function, and $I_{\mathrm{NCE}}$ serves as a variational lower bound of the mutual information $I$, with $N$ related to the batch size. This approach maximizes mutual information by estimating a lower bound.

This training process yields a language feature extraction model that preserves semantic structure while reducing spurious correlations in the embedding space. And we select the finetuned model with a batch size of 256 and 1000 training steps for this task. Some discussion on the performance of finetuning can be found in \cref{app:H-fine-tune-discussion}.

\subsection{Mitigation Methods Performance}\label{sec:3-5mitigation-methods-performance}

Our strategy effectively mitigates the issue in both \textit{text-based} (\cref{tab:table2}) and \textit{graph-based} settings (\cref{app:G-mitigation-result}). Experimental results show that in the \textit{text-based} scenario, our method achieves an average relative improvement of $\textbf{59\%}$ over the original model. In the \textit{graph-based} scenario, the average relative improvement reaches $\textbf{53\%}$. These results underscore the effectiveness of our strategy.

\section{Discussions}\label{sec:4discussion}

\subsection{Gap in the Exploration of Potential Shortcuts}\label{sec:4-1gap-in-the-exploration-of-potential-shortcut}

In our analysis, we observed a significant drop in model performance when the distribution of shallow textual features is altered. On the one hand, previous studies have shown that distributional biases in lexical usage, sentiment, and topic preference exist between social bot users and human users in commonly used datasets~\citep{breazeal2003emotion,li2025ets}. And by leveraging these biases, some approaches have achieved impressive performance in the social bot detection task. However, we argue that such performance gains are not inherently robust in our experiment.

\begin{figure}[t]
\hspace{0cm}
\includegraphics[width=1\linewidth]{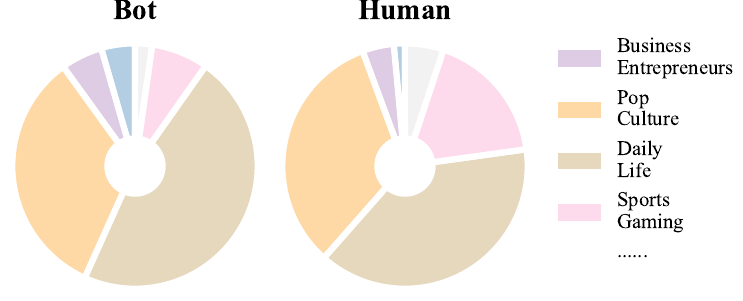}
\caption{
The topic distribution across different tags in the Cresci-2017-Data. There is no significant difference between human and bot distributions, indicating that topic features \emph{should not} be exploited as cues by detectors.}
\label{fig:fig_pie}
\end{figure}

On the other hand, we emphasize that models should not depend on these shortcuts, especially given that in real-world scenarios, the distributions of such features are often more balanced between user types. As shown in \cref{fig:fig_pie}, the topic distribution in benchmark dataset such as Cresci-2017-Data exhibits little divergence between humans and bots. This highlights the risk of shortcut learning and motivates the need for models that generalize beyond surface-level cues.

\subsection{Shortcut Scenarios Impact on Model}\label{sec:4-2shortcut-scenarios-impact-on-model}

In addition to demonstrating that shortcut scenarios significantly degrade model accuracy, we further examine how such features affect model confidence. Following the approach of~\citet{guo2017calibration}, we compute the expected calibration error (ECE) (the lower the better) and average prediction confidence.

We observe that when training stage occurs shortcut features, the model's accuracy drops, yet its prediction confidence increases notably in some cases. This suggests that the model becomes more overconfident despite making more errors, indicating a lack of proper uncertainty calibration in the presence of shortcut learning.

\begin{figure}[t]
\hspace{-0.7cm}
\includegraphics[width=1.1\linewidth]{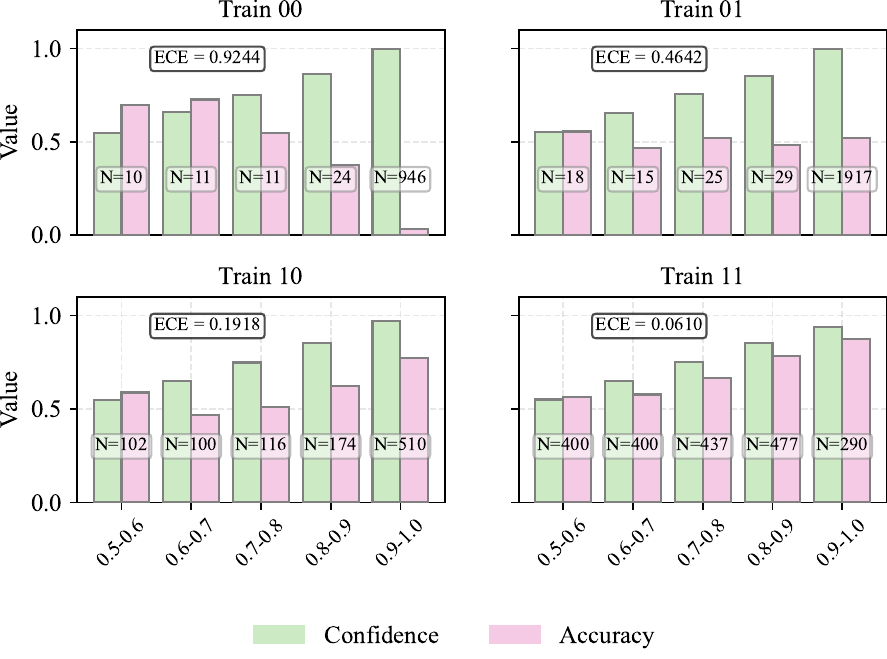}
\caption{Calibration of detectors in standard and shortcut settings. The sub-caption "Train $xy$" indicates that the model is trained under setting $x$ and tested under setting $y$, where 0 corresponds to the shortcut setting and 1 to the standard setting. The results demonstrate that models trained in the shortcut setting tend to exhibit higher confidence in their predictions, yet suffer from reduced accuracy.}
\label{fig:fig_ece}
\end{figure}

\subsection{The Appropriateness of Mitigation Strategies}\label{sec:4-3the-appropriateness-of-mitigation-strategies}

\begin{figure}[t]
\hspace{0cm}
\includegraphics[width=1\linewidth]{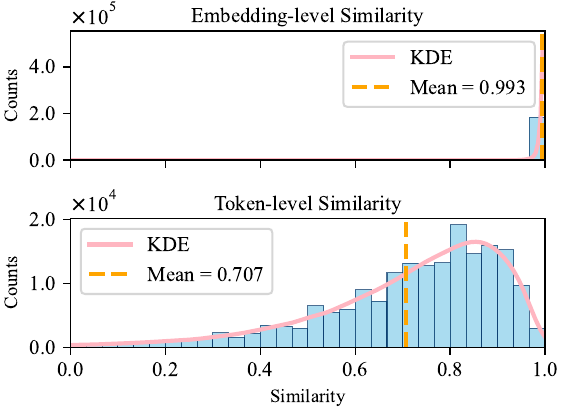}
\caption{Distributions of embedding-level and token-level similarity scores. Cosine similarities at embedding-level mostly exceed 0.9, and edit similarities at token-level (computed as $1-\frac{\text{edit distance}}{\text{raw tweet words}}$) are above 0.7 for the majority of text pairs, confirming that the rewriting process preserves semantic content while introducing minimal textual edits.}
\label{fig:sim}
\end{figure}

In prior work, \citet{feng2024does} demonstrated that LLMs can effectively distinguish between social bots and human users in detection tasks, thereby ensuring that our prompt-based rewriting preserves the original text labels. To further validate the effectiveness of LLMs in mitigating shortcut reliance, we evaluate the semantic consistency between the original and rewritten texts, confirming that only the targeted shallow features have been modified. We compute the edit distance at the token level and the cosine similarity at the embedding level between each pair of original and modified texts, and employ KDE to illustrate the overall distribution.(\cref{fig:sim}). The rewriting yields a token-level edit similarity of over 0.7 (substantially higher than the average similarity of 0.03 observed in over 1 million randomly paired texts), indicating that only a small portion of the original content is modified. And the embedding-level cosine similarity exceeds 0.9, demonstrating that the rewritten texts undergo minimal surface changes while maintaining high semantic consistency (see details and examples in \cref{app:E-LLMs-modify-example}).

Such minimal intervention supports the rationality of our mitigation strategies, which weaken shortcut patterns without distorting the core meaning and the underlying causal features relevant to classification.

\section{Related Work}\label{sec:5related-work}
%
\paragraph{Shortcut Learning.} Shortcut learning refers to the phenomenon where neural models exploit spurious correlations rather than truly understanding the underlying task~\citep{geirhos2020shortcut, du2023shortcut, wan2025truth}. This is a critical issue for ensuring the robustness of neural networks in real-world scenarios.
Previous studies have made significant contributions in robustness of social bot detection. For example, \citet{yang2020scalable} investigated how to utilize prototypical samples to improve the robustness of detection models. \citet{cresci2023demystifying} pointed out that some works exploit platform-specific spurious cues, such as verification status, to achieve better detection performance. \citet{hays2023simplistic} highlighted the challenges of generalization across datasets in social bot detection.
Recent research has also explored the use of dynamic information in the context of social bot detection \citep{zhou2023detecting, he2024botdgt}, or adopted more robust techniques such as unsupervised algorithms \citep{peng2024unsupervised} to enhance generalization performance.
Moreover, many recent studies have examined the increasing challenges of social bot detection in the era of LLMs \citep{ferrara2023social, feng2024does}, and some have examined how content generated by LLMs can trigger shortcut learning risks\citep{wan2025truth}. Collectively, these studies highlight the critical role of shortcut learning in the performance and generalization of social bot detection models.

\paragraph{Counterfactual Data Augmentation.} CDA seeks to improve model robustness by generating examples whose labels flip under a target classifier. Early efforts in counterfactual data augmentation relied on human-authored rewrites or contrast sets~\citep{kaushik2019learning,gardner2020evaluating}. Then various automated approaches have emerged, including rule-based perturbations~\citep{ribeiro2020beyond,webster2020measuring}, control-guided generation~\citep{madaan2021generate,wu2021polyjuice,ross2021tailor}, and retrieval-based methods leveraging external knowledge~\citep{paranjape2021retrieval}. Recent work explores explanation-guided and distillation-based methods~\citep{kim2023grounding,jeanneret2024text}. With the rise of LLMs, prompt-based counterfactual generation has become increasingly popular~\citep{madaan2023self,chiang2023can,zheng2023judging,kocmi2023large,mishra2024llm}. We use LLMs to generate feature-specific counterfactual augmentation for capturing causal feature in social bot detection.

\section{Conclusion}\label{sec:6conculsion}

Our work reveals that shortcut features significantly affect social bot detection performance across multiple textual feature dimensions through triggering shortcut learning in training process. Building on counterfactual data augmentation using LLMs, we explore their effectiveness in addressing this challenging distribution shift from both the data and feature extraction perspectives. This work highlights the robustness gains achieved through this approach and offer new insights into designing more resilient social bot detectors. Moreover, our study provides a novel perspective on leveraging LLMs for causal interventions in scenarios with unknown or implicit ground truth. To the best of our knowledge, this is the first in-depth investigation into the impact of shortcut learning and the capacity for causal information extraction on social bot detection.

\section*{Acknowledgments}
This work is supported in part by the National Natural Science
Foundation of China (No. 62192781, No. 62272374),
the Natural Science Foundation of Shaanxi Province (No.
2024JC-JCQN-62), the State Key Laboratory of Communication
Content Cognition under Grant No. A202502, the
Key Research and Development Project in Shaanxi Province
(No. 2023GXLH-024), the Project of China Knowledge
Center for Engineering Science and Technology, Huawei-Xi' an Jiaotong
University Elite Class Program (No. INCCHN2508012229)
and the K. C. Wong Education Foundation.

\bibliography{aaai2026}

\makeatletter
\let\orig@listi\@listi
\let\orig@listii\@listii
\let\orig@listiii\@listiii
\let\origparagraph\paragraph
\let\origthesubsection\thesubsection
\let\origlabelenumi\labelenumi


\clearpage 
\appendix
\section{Limitation}\label{app:A-limitation}
While this study primarily focuses on revealing the vulnerability of social bot detectors to shortcut learning and proposes practical mitigation strategies, there are certain limitations that should be acknowledged. Our analysis centers on textual features, as text remains the most dominant modality in existing benchmark datasets. Although we did not include more recent state-of-the-art (SOTA) models with advanced feature extraction capabilities, it is worth noting that many of these SOTA models build upon the same baseline architectures we adopt in this work. As such, our findings remain highly relevant and transferable. Furthermore, the scenarios we designed are derived from benchmark datasets, which may not fully capture the complexity of real-world environments. Nevertheless, we believe these controlled settings provide valuable insights and a solid foundation for future work to explore more diverse and realistic scenarios.

\section{Social Bot Detection Background}\label{app:B-social-bot-detection-background}

\paragraph{Social bot detection.} Social bot detection is a topic of great practical importance and has been extensively studied due to its profound influence on online ecosystems. The inherent ambiguity in task definition, coupled with the variability of application scenarios, makes this a particularly challenging research area. 

Most of the current social bot detection methods typically rely on large-scale training data and are designed to extract diverse user and network features. These approaches have demonstrated impressive performance on benchmark datasets and have significantly advanced the understanding of bot behaviors~\citep{liu2023botmoe}. Existing methods are commonly categorized into feature-based, text-based, and graph-based approaches. 
Feature-based methods utilize handcrafted features derived from user metadata and tweet content ~\citep{hays2023simplistic, wu2023botshape}, combined with classifiers such as SMOTEN N~\citep{kudugunta2018deep} and anomaly detection techniques~\citep{chavoshi2016debot}.
Text-based approaches leverage natural language processing models, including LSTMs~\citep{luo2020deepbot}, attention mechanisms~\citep{feng2021satar}, and pretrained transformers~\citep{dukic2020you, feng2024does}, to analyze user-generated content~\citep{cai2024lmbot}.
Graph-based methods focus on the structural characteristics of social networks~\citep{dehghan2023detecting}, employing techniques such as centrality measures and heterogeneous graph neural networks to model user interactions and relationships~\citep{ali2019detect, feng2021botrgcn}. 

Some efforts also explore dynamic graph~\citep{he2024botdgt} structures, graph entropy~\citep{yang2024sebot} and some online methods~\citep{zhou2023detecting} to enhance detection performance and robustness. Focusing on the robustness of social bot detection, this paper addresses the unexplored gap of shortcut learning in social bot detection.

\section{Detailed Scenario Design}\label{app:C-detailed-scenario-design}
Here, we provide a more detailed description of the potential shortcut setup.
Firstly, we provide a detailed explanation of how we define the $\mathcal{U}^{\text{pos}}_{\text{fea}}$ and $\mathcal{U}^{\text{neg}}_{\text{fea}}$ subsets for each shallow feature, namely, sentiment, topic, emotion, and human values.

\textbf{1. Sentiment:}  
In exploring this shallow feature, we primarily focus on the impact of shortcut learning driven by sentiment polarity (\emph{e.g.}, positive or negative tone) on model performance. We use the sentiment classifier model, \textit{Cardiffnlp/Twitter-RoBERTa-base-sentiment-latest}~\citep{camacho-collados-etal-2022-tweetnlp,loureiro-etal-2022-timelms} to extract sentiment from the text. We define:
\begin{align*}
\mathcal{U}^{\text{pos}}_{\text{senti}} &= \{ \text{user} \mid \text{user text exhibits} \ \textit{positive} \ \text{sentiment}\},\\
\mathcal{U}^{\text{neg}}_{\text{senti}} &= \{ \text{user} \mid \text{user text exhibits} \ \textit{negative} \ \text{sentiment}\}.
\end{align*}

\textbf{2. Topic:}  
In analyzing this shallow feature, we investigate the influence of shortcut learning induced by different topic styles on model performance. We divide topic into "dailylife-related" and "Non-daliylife-related" categories, and set up shortcuts based on these. We use a topic analysis model, \textit{Cardiffnlp/Twitter-RoBERTa-base-dec2021-tweet-topic-single-all}~\citep{camacho-collados-etal-2022-tweetnlp,loureiro-etal-2022-timelms} to extract topic from the text, defined as:
\begin{align*}
\mathcal{U}^{\text{pos}}_{\text{topic}} &= \left\{
\text{user} 
\begin{array}{l}
\mid \text{user text discusses the topic of} \ \textit{daily life} \end{array}\right\},\\
\mathcal{U}^{\text{neg}}_{\text{topic}} &= \left\{ \text{user} \mid 
\begin{array}{l}
\text{user text discusses the topic of} \\ \textit{pop culture, sports} 
\end{array}
\right\}.
\end{align*}

\textbf{3. Emotion:}  
While analyzing this superficial feature, we focus on the impact of shortcut learning generated by different emotional styles on model performance. We divide emotions into two different categories and set up shortcuts accordingly. We use an emotion analysis model, \textit{Cardiffnlp/Twitter-RoBERTa-large-emotion-latest}~\citep{antypas2023supertweeteval} to extract emotions from the text, and the scenario is defined as:
\begin{align*}
\mathcal{U}^{\text{pos}}_{\text{emotion}} &= \left\{ \text{user} \mid \begin{array}{l}
\text{user text expresses emotions of} \\ \textit{anticipation, optimism}
\end{array} \right\},\\
\mathcal{U}^{\text{neg}}_{\text{emotion}} &= \{ \text{user} \mid \text{user text expresses emotions of}\ \textit{joy}\}.
\end{align*}

\textbf{4. Human Values:}  
In examining this surface-level attribute, we focus on the impact of shortcut learning generated by different value orientations on model performance. We divide values into "Supportive" and "Oppositional" categories, and set up shortcuts accordingly. We use human values extracting model, \textit{VictorYeste/DeBERTa-based-human-value-detection}~\citep{yeste2024philo} to extract values from the text, and define $\mathcal{U}^{\text{pos}}_{\text{value}}$, $\mathcal{U}^{\text{neg}}_{\text{value}}$ as:
\begin{align*}
\mathcal{U}^{\text{pos}}_{\text{value}} &= \left\{ \text{user} \mid \begin{array}{l}
\text{users whose text values belong to} \\
\textit{Face, Hedonism, Achievement, }\\
\textit{Security: personal}
\end{array} \right\},\\
\mathcal{U}^{\text{neg}}_{\text{value}} &= \left\{ \text{user} \mid \begin{array}{l}
\text{users whose text values belong to} \\
\textit{Self-direction: thought},\\
\textit{Conformity: interpersonal, }\\
\textit{Security: societal}
\end{array} \right\}.
\end{align*}

Secondly, we present a systematic and transparent process for constructing the dataset.

For each dataset, we follow a unified processing pipeline consisting of four main steps: dataset cleaning, text feature extraction, user partitioning, and scenario dataset splitting.
First, we clean the raw dataset into a standardized format to facilitate downstream processing. A key step here is treating tweets as user-level features.
Then, we extract shallow textual features from the tweets across different dimensions (\emph{e.g.}, sentiment, topic, emotion, value) using specialized LLMs.
Next, we divide users into subsets according to the definitions of $\mathcal{U}^{\text{pos}}_{\text{fea}}$, $\mathcal{U}^{\text{neg}}_{\text{fea}}$, and $\mathcal{U}^{\text{neu}}_{\text{fea}}$ introduced in the previous section. We select representative tweets for each user that reflect the corresponding attribute. Users are then grouped into six categories per feature dimension: $\text{Bot}^{\text{pos}}_{\text{fea}}$, $\text{Bot}^{\text{neg}}_{\text{fea}}$, $\text{Bot}^{\text{neu}}_{\text{fea}}$, $\text{Human}^{\text{pos}}_{\text{fea}}$, $\text{Human}^{\text{neg}}_{\text{fea}}$, and $\text{Human}^{\text{neu}}_{\text{fea}}$.
Finally, we construct the shortcut learning setting based on these groups. To eliminate confounding factors, we balance the number of samples in the training and test sets across both classes with a 1:1 ratio and the validation set is split from the training set (See \cref{tab:set_num} for dataset partitioning details).

\begin{table}[t]
\centering
\begin{tabular}{lll|cc}
\toprule
\multicolumn{3}{c}{Sample Number of Set} & Standard & Shortcut \\
\cmidrule (lr){1-5}
\multirow{8}{*}{{\small Cresci-2015-Data}} & \multirow{2}{*}{Sentiments} & Train & 1524 & 1526 \\
 &  & Test & 764 & 382 \\
\cmidrule (lr){2-5}
 & \multirow{2}{*}{Emotions} & Train & 1552 & 1552 \\
 &  & Test & 780 & 390 \\
\cmidrule (lr){2-5}
 & \multirow{2}{*}{Topics} & Train & 1556 & 1556 \\
 &  & Test & 780 & 390 \\
\cmidrule (lr){2-5}
 & \multirow{2}{*}{Values} & Train & 1552 & 1554 \\
 &  & Test & 780 & 390 \\
\cmidrule (lr){1-5}
\multirow{8}{*}{{\small Cresci-2017-Data}} & \multirow{2}{*}{Sentiments} & Train & 864 & 864 \\
 &  & Test & 432 & 216 \\
\cmidrule (lr){2-5}
 & \multirow{2}{*}{Emotions} & Train & 864 & 864 \\
 &  & Test & 436 & 218 \\
\cmidrule (lr){2-5}
 & \multirow{2}{*}{Topics} & Train & 864 & 864 \\
 &  & Test & 436 & 218 \\
\cmidrule (lr){2-5}
 & \multirow{2}{*}{Values} & Train & 864 & 864 \\
 &  & Test & 436 & 218 \\
\cmidrule (lr){1-5}
\multirow{8}{*}{{\small Twibot-20}} & \multirow{2}{*}{Sentiments} & Train & 4004 & 4004 \\
 &  & Test & 2004 & 1002 \\
\cmidrule (lr){2-5}
 & \multirow{2}{*}{Emotions} & Train & 4128 & 4128 \\
 &  & Test & 2064 & 1032 \\
\cmidrule (lr){2-5}
 & \multirow{2}{*}{Topics} & Train & 4140 & 4142 \\
 &  & Test & 2072 & 1036 \\
\cmidrule (lr){2-5}
 & \multirow{2}{*}{Values} & Train & 4080 & 4082 \\
 &  & Test & 2044 & 1022 \\
\bottomrule
\end{tabular}
\caption{
Dataset partition statistics. The combination of training and testing settings, standard or shortcut, indicates whether shortcut features were introduced during training or appear in the test set. Sample counts for each scenario are reported, with bots and humans balanced at a 1:1 ratio, demonstrating that our setting is sufficient and objective. (Note that the validation set is split from the training set.)
}
\label{tab:set_num}
\end{table}

\begin{table*}[t]
\centering
\begin{tabular}{ll|cc|cc}
\toprule
\multicolumn{2}{c}{\multirow{2}{*}{BotRGCN}} & \multicolumn{2}{c}{{\small Cresci-2015-Data}} & \multicolumn{2}{c}{{\small Twibot-20}} \\
 &  & {$\text{\small Shortcut}_{\text{te}}$} & {$\text{\small Standard}_{\text{te}}$} & {$\text{\small Shortcut}_{\text{te}}$} & {$\text{\small Standard}_{\text{te}}$} \\
\cmidrule(lr){1-6}
\multirow{3}{*}{Sentiments} & $\text{Standard}_{\text{tr}}$ & .961 & .971 & .710 & .702 \\
 & $\text{Shortcut}_{\text{tr}}$ & .749 & .869 & .060 & .522 \\
 & Difference & 22\%$\downarrow$ & 10\%$\downarrow$ & 91\%$\downarrow$ & 25\%$\downarrow$ \\
\cmidrule(lr){1-6}
\multirow{3}{*}{Emotions} & $\text{Standard}_{\text{tr}}$ & .985 & .979 & .714 & .717 \\
 & $\text{Shortcut}_{\text{tr}}$ & .903 & .941 & .126 & .539 \\
 & Difference & 8\%$\downarrow$ & 3\%$\downarrow$ & 82\%$\downarrow$ & 24\%$\downarrow$ \\
\cmidrule(lr){1-6}
\multirow{3}{*}{Topics} & $\text{Standard}_{\text{tr}}$ & .987 & .973 & .732 & .731 \\
 & $\text{Shortcut}_{\text{tr}}$ & .967 & .968 & .187 & .548 \\
 & Difference & 2\%$\downarrow$ & 0\%$\downarrow$ & 74\%$\downarrow$ & 25\%$\downarrow$ \\
\cmidrule(lr){1-6}
\multirow{3}{*}{Values} & $\text{Standard}_{\text{tr}}$ & .926 & .951 & .745 & .725 \\
 & $\text{Shortcut}_{\text{tr}}$ & .815 & .904 & .215 & .564 \\
 & Difference & 11\%$\downarrow$ & 4\%$\downarrow$ & 71\%$\downarrow$ & 22\%$\downarrow$ \\
\bottomrule
\end{tabular}
\caption{Accuracy of BotRGCN model under standard and shortcut training and testing conditions. The vertical dimension denotes training configurations, $\text{Standard}_{\text{tr}}$ (randomly sampled, balanced labels) versus $\text{Shortcut}_{\text{tr}}$ (spuriously correlated shallow features), while the horizontal dimension denotes test feature distributions, $\text{Standard}_{\text{te}}$ (balanced) versus $\text{Shortcut}_{\text{te}}$ (distributional shift). The difference between paired entries indicates the drop in accuracy when moving from standard to shortcut settings, reflects the model’s vulnerability to spurious correlations.}
\label{tex:table_rgcn_shortcut}
\end{table*}

\begin{table*}[t]
\centering
\begin{tabular}{ll|cc|cc|cc}
\toprule
\multicolumn{2}{c}{\multirow{2}{*}{AMR+CIGA}} & \multicolumn{2}{c}{{\small Cresci-2015-Data}} & \multicolumn{2}{c}{{\small Cresci-2017-Data}} & \multicolumn{2}{c}{{\small Twibot-20}} \\
 &  & {$\text{\small Shortcut}_{\text{te}}$} & {$\text{\small Standard}_{\text{te}}$} & {$\text{\small Shortcut}_{\text{te}}$} & {$\text{\small Standard}_{\text{te}}$} & {$\text{\small Shortcut}_{\text{te}}$} & {$\text{\small Standard}_{\text{te}}$} \\
\cmidrule(lr){1-8}
\multirow{3}{*}{Sentiments} & $\text{Standard}_{\text{tr}}$ & .873 & .917 & .734 & .773 & .569 & .567 \\
 & $\text{Shortcut}_{\text{tr}}$ & .678 & .838 & .271 & .671 & .112 & .531 \\
 & Difference & 22\%$\downarrow$ & 8\%$\downarrow$ & 63\%$\downarrow$ & 13\%$\downarrow$ & 80\%$\downarrow$ & 6\%$\downarrow$ \\
\cmidrule(lr){1-8}
\multirow{3}{*}{Emotions} & $\text{Standard}_{\text{tr}}$ & .948 & .938 & .806 & .830 & .606 & .582 \\
 & $\text{Shortcut}_{\text{tr}}$ & .770 & .870 & .382 & .679 & .232 & .538 \\
 & Difference & 18\%$\downarrow$ & 7\%$\downarrow$ & 52\%$\downarrow$ & 18\%$\downarrow$ & 61\%$\downarrow$ & 7\%$\downarrow$ \\
\cmidrule(lr){1-8}
\multirow{3}{*}{Topics} & $\text{Standard}_{\text{tr}}$ & .896 & .904 & .806 & .773 & .612 & .556 \\
 & $\text{Shortcut}_{\text{tr}}$ & .774 & .850 & .405 & .656 & .189 & .530 \\
 & Difference & 13\%$\downarrow$ & 5\%$\downarrow$ & 49\%$\downarrow$ & 15\%$\downarrow$ & 69\%$\downarrow$ & 4\%$\downarrow$ \\
\cmidrule(lr){1-8}
\multirow{3}{*}{Values} & $\text{Standard}_{\text{tr}}$ & .797 & .858 & .776 & .802 & .579 & .596 \\
 & $\text{Shortcut}_{\text{tr}}$ & .592 & .785 & .572 & .739 & .225 & .536 \\
 & Difference & 25\%$\downarrow$ & 8\%$\downarrow$ & 26\%$\downarrow$ & 7\%$\downarrow$ & 61\%$\downarrow$ & 10\%$\downarrow$ \\
\bottomrule
\end{tabular}
\caption{Accuracy of AMR+CIGA model under standard and shortcut training and testing conditions. The vertical dimension denotes training configurations, $\text{Standard}_{\text{tr}}$ (randomly sampled, balanced labels) versus $\text{Shortcut}_{\text{tr}}$ (spuriously correlated shallow features), while the horizontal dimension denotes test feature distributions, $\text{Standard}_{\text{te}}$ (balanced) versus $\text{Shortcut}_{\text{te}}$ (distributional shift). The difference between paired entries indicates the drop in accuracy when moving from standard to shortcut settings, reflects the model’s vulnerability to spurious correlations.}
\label{tex:table_ciga}
\end{table*}

\section{Potential Shortcut}\label{app:D-potential-shortcut}
\Cref{tex:table_rgcn_shortcut,tex:table_ciga} present additional experimental results on shortcut scenarios in graph-based models and basic debiasing methods. The results demonstrate performance degradation across different settings and reveal the limitations of conventional debiasing approaches in addressing such cases. 
In the graph-based method, compared to the normal distribution scenario, the relative classification performance in \textit{shortcut test setting} on average decreased by $\textbf{45\%}$, with the largest relative decrease being $\textbf{91\%}$. In \textit{standard test setting}, the average relative performance drop was $\textbf{15\%}$, with the most significant relative decrease being $\textbf{25\%}$. And in the AMR+CIGA method, compared to the normal distribution scenario, the classification performance in \textit{shortcut test setting} on average decreased by $\textbf{45\%}$, with the largest decrease being $\textbf{80\%}$. In \textit{standard test setting}, the average performance drop was $\textbf{9\%}$, with the most significant decrease being $\textbf{18\%}$.

Additionally, in our AMR+CIGA implementation, we leverage the graph structure to facilitate feature disentanglement and employ mature causal‐feature extraction methods. First, we convert each input sentence into an Abstract Meaning Representation (AMR) graph using the model-parse-t5-v0-2.0 parser. Next, we apply the CIGA framework~\citep{chen2022learning} directly on these graph‐structured texts. By combining AMR’s explicit semantic graphs with CIGA’s causal disentanglement, we explore a broadly adopted approach to bias mitigation through feature disentanglement in this setting.

\begin{table*}[t]
\centering
\begin{tabular}{c|p{8cm}|p{3cm}}
\toprule
Emotion & \multicolumn{1}{c}{Text} & \multicolumn{1}{c}{Feature} \\
\hhline{===}
Raw Tweets & (stasera ne ho fatte \textbf{arrabbiare} due, e se servisse ne ho altre due \textbf{arrabbiate} da settimane di riserva) & \multicolumn{1}{c}{Anticipation} \\
\cmidrule (lr){1-3}
Re Write & (stasera ne ho fatte \textbf{ridere} due, e se servisse ne ho altre due \textbf{felici} da settimane di riserva) & \multicolumn{1}{c}{Joy} \\
\hhline{===}
Raw Tweets & RT @nbccommunity: ONE MONTH! \textbf{\#ThatsOctober19th} & \multicolumn{1}{c}{Anticipation} \\
\cmidrule (lr){1-3}
Re Write & RT @nbccommunity: ONE MONTH! \textbf{\#SoExcitedForOctober19th} & \multicolumn{1}{c}{Joy} \\
\hhline{===}
Raw Tweets & Decisi che ero \textbf{bella} per il semplice motivo che avevo voglia di esserlo. \{Eva Luna di Isabel Allende\} & \multicolumn{1}{c}{Anticipation} \\
\cmidrule (lr){1-3}
Re Write & Decisi che ero \textbf{felice} per il semplice motivo che avevo voglia di esserlo. \{Eva Luna di Isabel Allende\} & \multicolumn{1}{c}{Joy} \\
\hhline{===}
Raw Tweets & @ValentinaBorgo si quelli che mi defollowano sono gli stessi che credono che twitter sia uguale a facebook, \textbf{probabilmente ;)} & \multicolumn{1}{c}{Anticipation} \\
\cmidrule (lr){1-3}
Re Write & @ValentinaBorgo si quelli che mi defollowano sono gli stessi che credono che twitter sia uguale a facebook, \textbf{fantasticamente!)} & \multicolumn{1}{c}{Joy} \\
\hhline{===}
Raw Tweets & RT @lapopistelli: Dati quasi definitivi. Hanno votato 3.100.000 circa e Bersani \textbf{va al ballottaggio} con 10.2 punti di vantaggio & \multicolumn{1}{c}{Joy} \\
\cmidrule (lr){1-3}
Re Write & RT @lapopistelli: Dati quasi definitivi. Hanno votato 3.100.000 circa e Bersani \textbf{potrebbe vincere} con 10.2 punti di vantaggio & \multicolumn{1}{c}{Anticipation} \\
\bottomrule
\end{tabular}
\caption{We present an example of large language model rewriting. The results of counterfactual data augmentation on the emotion textual attribute are shown, with differences between the original and rewritten texts highlighted in bold. The minimal nature of these changes demonstrates the precision of the modification, validating the effectiveness of our counterfactual augmentation approach in ensuring that the model learns emotion-independent cues for social bot detection.}
\label{tex:table_emotion}
\end{table*}

\begin{table*}[t]
\centering
\begin{tabular}{c|p{8cm}|p{3cm}}
\toprule
Sentiment & \multicolumn{1}{c}{Text} & \multicolumn{1}{c}{Feature} \\
\hhline{===}
Raw Tweets & Having a \textbf{hard} time with heaps \& priority queues & \multicolumn{1}{c}{Negative} \\
\cmidrule (lr){1-3}
Re Write & Having a \textbf{great} time with heaps \& priority queues & \multicolumn{1}{c}{Positive} \\
\hhline{===}
Raw Tweets & @Chris Sixx looks \textbf{nice!} but no push & \multicolumn{1}{c}{Positive} \\
\cmidrule (lr){1-3}
Re Write & @Chris Sixx looks \textbf{awful!} but no push & \multicolumn{1}{c}{Negative} \\
\hhline{===}
Raw Tweets & @intervistato @jacopopaoletti @sednonsatiata \textbf{ok :)} intanto il mio linkedin lo trovate nei miei tweet di ieri. & \multicolumn{1}{c}{Positive} \\
\cmidrule (lr){1-3}
Re Write & @intervistato @jacopopaoletti @sednonsatiata \textbf{ugh : (} intanto il mio linkedin lo trovate nei miei tweet di ieri. & \multicolumn{1}{c}{Negative} \\
\hhline{===}
Raw Tweets & @occhidaoriental ma te pare:D e a te un bel maglioncino a righe blue elettrico e giallo evidenziatore con collana verde pisello non \textbf{serve?:D} & \multicolumn{1}{c}{Positive} \\
\cmidrule (lr){1-3}
Re Write & @occhidaoriental ma te pare:D e a te un bel maglioncino a righe blue elettrico e giallo evidenziatore con collana verde pisello non \textbf{serve?: (} & \multicolumn{1}{c}{Neutral} \\
\hhline{===}
Raw Tweets & RT @charlysisto: @jseifer \textbf{strangely} my railsconf proposal "ruby on berlusconi" was \textbf{not} accepted \#railsconf & \multicolumn{1}{c}{Negative} \\
\cmidrule (lr){1-3}
Re Write & RT @charlysisto: @jseifer \textbf{amazingly} my railsconf proposal "ruby on berlusconi" was \textbf{enthusiastically} accepted \#railsconf & \multicolumn{1}{c}{Positive} \\
\bottomrule
\end{tabular}
\caption{We present an example of large language model rewriting. The results of counterfactual data augmentation on the sentiment textual attribute are shown, with differences between the original and rewritten texts highlighted in bold. The minimal nature of these changes demonstrates the precision of the modification, validating the effectiveness of our counterfactual augmentation approach in ensuring that the model learns sentiment-independent cues for social bot detection.}
\label{tex:table_sentiment}
\end{table*}

\begin{table*}[t]
\centering
\begin{tabular}{c|p{8cm}|p{3cm}}
\toprule
Topic & \multicolumn{1}{c}{Text} & \multicolumn{1}{c}{Feature} \\
\hhline{===}
Raw Tweets & @liz nicole Congratulations on your new \textbf{title,} then! :) & \multicolumn{1}{c}{Pop culture} \\
\cmidrule (lr){1-3}
Re Write & @liz nicole Congratulations on your new \textbf{job,} then! :) & \multicolumn{1}{c}{Daily life} \\
\hhline{===}
Raw Tweets & RT @maboa: Note to self : when it's raining, \textbf{code} to jazz. & \multicolumn{1}{c}{Pop culture} \\
\cmidrule (lr){1-3}
Re Write & RT @maboa: Note to self : when it's raining, \textbf{walk} to jazz. & \multicolumn{1}{c}{Daily life} \\
\hhline{===}
Raw Tweets & @kekkoz ah vedi. maledetto \textbf{mymovies} che non lo specifica. & \multicolumn{1}{c}{Pop culture} \\
\cmidrule (lr){1-3}
Re Write & @kekkoz ah vedi. maledetto \textbf{supermercato} che non lo specifica. & \multicolumn{1}{c}{Daily life} \\
\hhline{===}
Raw Tweets & @JackGrifo @kikkalenzi @79 fra @ANFAMEEE @per piacere ma sono cose che non dovrebbero uscire dallo \textbf{spogliatoio} :-) & \multicolumn{1}{c}{Daily life} \\
\cmidrule (lr){1-3}
Re Write & @JackGrifo @kikkalenzi @79 fra @ANFAMEEE @per piacere ma sono cose che non dovrebbero uscire dallo \textbf{stadio} :-) & \multicolumn{1}{c}{Pop culture} \\
\hhline{===}
Raw Tweets & @mlovesociety I'm flying to \textbf{Tegel} for \textbf{\#mlove} on 27th afternoon and heading back to \textbf{Berlin} on sat 30th morning any \textbf{carpooling} sharing op? & \multicolumn{1}{c}{Daily life} \\
\cmidrule (lr){1-3}
Re Write & @mlovesociety I'm flying to \textbf{LA} for \textbf{\#Coachella} on 27th afternoon and heading back to \textbf{Vegas} on sat 30th morning any \textbf{ride} sharing op? & \multicolumn{1}{c}{Pop culture} \\
\bottomrule
\end{tabular}
\caption{Example of large language model rewriting. Results of counterfactual data augmentation on the topic textual attribute are shown, highlighting differences between the original and rewritten texts in bold. The changes are minimal, demonstrating precise modification.}
\label{tex:table_topic}
\end{table*}

\begin{table*}[t]
\centering
\begin{tabular}{c|p{8cm}|p{3cm}}
\toprule
Value & \multicolumn{1}{c}{Text} & \multicolumn{1}{c}{Feature} \\
\hhline{===}
Raw Tweets & @dzovan same to you, \textbf{storm} permitting :) & \multicolumn{1}{c}{Benevolence: caring} \\
\cmidrule (lr){1-3}
Re Write & @dzovan same to you, \textbf{party} permitting :) & \multicolumn{1}{c}{Conformity: rules} \\
\hhline{===}
Raw Tweets & @bwindham you would \textbf{love} this- all the neighborhood children are \textbf{angels...} & \multicolumn{1}{c}{Benevolence: caring} \\
\cmidrule (lr){1-3}
Re Write & @bwindham you would \textbf{adore} this- all the neighborhood children are \textbf{partying...} & \multicolumn{1}{c}{Hedonism} \\
\hhline{===}
Raw Tweets & @thiswildidea so my pups NYC rescue but we live in Rome. Can we still \textbf{participate?} & \multicolumn{1}{c}{Universalism: concern} \\
\cmidrule (lr){1-3}
Re Write & @thiswildidea so my pups NYC rescue but we live in Rome. Can we still \textbf{party?} & \multicolumn{1}{c}{Hedonism} \\
\hhline{===}
Raw Tweets & @Ettoreeeee lo si fa per \textbf{ridere} un po' :) & \multicolumn{1}{c}{Benevolence: caring} \\
\cmidrule (lr){1-3}
Re Write & @Ettoreeeee lo si fa per \textbf{divertirsi} un po' :) & \multicolumn{1}{c}{Face} \\
\hhline{===}
Raw Tweets & Now we are at YSL homme, rue Faubourg St. Honore' & \multicolumn{1}{c}{Universalism: concern} \\
\cmidrule (lr){1-3}
Re Write & Now we are at YSL homme, \textbf{sipping champagne} rue Faubourg St. Honore' & \multicolumn{1}{c}{Stimulation} \\
\bottomrule
\end{tabular}
\caption{Example of large language model rewriting. Results of counterfactual data augmentation on the value textual attribute are shown, highlighting differences between the original and rewritten texts in bold. The changes are minimal, demonstrating precise modification.}
\label{tex:table_value}
\end{table*}

\section{LLMs Modify Examples}\label{app:E-LLMs-modify-example}

We provide several examples of LLM-rewritten tweets from our experiments using the DeepSeek API (\cref{tex:table_emotion,tex:table_sentiment,tex:table_topic,tex:table_value}). As illustrated in the examples, the rewritten tweets differ only slightly from the originals, demonstrating that the LLMs perform fine-grained modifications while preserving core semantics. And we further illustrate that the changes between text pairs are minimal, both in terms of embedding similarity and word-level similarity. From embedding similarity perspective, we compute cosine similarity between each pair, achieving an average value of 0.993, which indicates strong semantic consistency. And in word level, we calculate the normalized word-level edit similarity, defined as $1 - \frac{D(w_1, w_2)}{|w_1|}$, where $D(w_1, w_2)$ is the Levenshtein distance in word-level between text. This yields an average similarity of 0.707, which is substantially higher than the average of 0.03 observed in random text pairs based on one million samples.

In addition, the interaction with the large language model is conducted using a standardized prompting template, which ensures consistency and controllability across all rewriting tasks. The template is detailed as follows. For simplicity and consistency, we use the label-free prompt in practice for all data.

\begin{tcolorbox}[colback=gray!5!white, colframe=gray!80!black, title=Prompt Template Used for LLM Rewriting]

This tweet shows \{attribute type\} \{attribute\} and the user is a bot or human. Please change as few words as possible to rewrite it to different \{attribute\} and do not change the sentence structure to keep the user's feature. Your answer just contain output without any extra content.

Input: [\{text to modify\}]

Output:,
\end{tcolorbox}

\begin{quote}
\textbf{System:} "You are an expert in social network analysis, focusing on the detection of social bots. You are particularly interested in addressing distribution biases in shallow textual features in topic between bots and human users. Your goal is to modify or augment the text to mitigate this bias while preserving the original semantics."

\textbf{User:} "This tweet's user is a bot or human, this tweets shows pop culture. Please change as few words as possible to rewrite it to different topics and don't change the sentence structure to keep the human feature. your answer just contain output without any extra content.\\
Input: ['RT @NikkiFinke: Average Movie Ticket Price Falls To \$7.78 [URL] via @DeadlineCom']\\
Output:,"

\textbf{Bot:}"RT @DailyNews: Average Grocery Bill Falls To \$78.50 [URL] via @LocalMarket"

\textbf{User:} "This tweet's user is a bot or human, this tweets shows optimism. Please change as few words as possible to rewrite it to different emotions and don't change the sentence structure to keep the user's feature. your answer just contain output without any extra content.\\
Input: ["Narrative was the elegant Trojan horse out of which the most daring and radical ideas could explode and storm a great city -@tanehisicoates]\\
Output:,"

\textbf{Bot:} "Narrative was the delightful rainbow bridge over which the most cheerful and uplifting ideas could dance and brighten a great city-@tanehisicoates"

\textbf{User:} \textbf{$\cdots$}

\textbf{Bot:} \textbf{$\cdots$}
\end{quote}

\begin{algorithm}[tb]
\caption{Text-Level Debiasing Strategy}
\label{alg:algorithm1}
\textbf{Input}: 
Raw tweets $\mathcal{T} = \{T_1,\dots,T_k\}$ and rewritten tweets $\mathcal{T'}=\{T'_1,\dots,T'_k\}$ of a user\\
\textbf{Output}: Debiased tweet set $\mathcal{T''}$ of a user\\
\begin{algorithmic}[1]
\STATE Initialize $\text{best\_score}\gets \infty$, $\text{best\_sel}\gets \mathbf{0}^k$
\FOR{each binary selection vector $S\in\{0,1\}^k$}
    \STATE Build candidate set 
    $$T^S_i \;=\; \begin{cases}
      T_i, & S_i=0\\
      T'_i, & S_i=1
    \end{cases}
    \quad\text{for }i=1,\dots,k$$
    \STATE Compute bias score $f$ over $T^S$ using \cref{eq:balance_score}
    \IF{$f < \text{best\_score}$}
        \STATE $\text{best\_score}\gets f$, \quad $\text{best\_sel}\gets S$
    \ENDIF
\ENDFOR
\STATE Construct final debiased set 
$$T''_i \;=\; \begin{cases}
    T_i, & \text{if } \text{best\_sel}_i=0,\\
    T'_i, & \text{if } \text{best\_sel}_i=1
\end{cases}
\quad\forall i$$
\STATE \textbf{return} $\mathcal{T''}$
\end{algorithmic}
\end{algorithm}

\section{Algorithms}\label{app:F-algorithms}

Here we show the pseudocode for our mitigation at text-level in \cref{alg:algorithm1} and dataset-level in \cref{alg:algorithm2}.

\begin{algorithm}[tb]
\caption{Dataset-Level Debiasing}
\label{alg:algorithm2}
\textbf{Input}: Training set $\mathcal{D} = \{ ( (\mathcal{T},\mathcal{T'},\mathcal{T''}), y)_{\text{user}_i}\}^N$\\
\textbf{Output}: Debiased training set $\mathcal{D}'$\\

\begin{algorithmic}[1]
\STATE Initialize $\mathcal{D}' \gets \emptyset$
\FOR{each label class $y \in \{0,1\}$}
    \STATE $\mathcal{U} \gets \{\text{user}_i \mid y_{\text{user}_i} = y\}$ \COMMENT{Indices of samples with label $y$}
    \STATE Shuffle $\mathcal{U}$
    \STATE $m \gets |\mathcal{U}|$
    \STATE $h \gets \lfloor m/2\rfloor$
    \STATE $\mathcal{U}_{\text{orig}} \gets \mathcal{U}[\ :h]$\quad\COMMENT{Assign original tweets $\mathcal{T}$}
    \STATE $\mathcal{U}_{\text{rew}} \gets \mathcal{U}[h+1:\ ]$\quad\COMMENT{Assign rewritten tweets $\mathcal{T'}$}
    \FOR{each $\text{user}_i \in \mathcal{U}_{\text{rew}}$}
        \IF{$\mathcal{T'}_{user_i}$ is valid }
            \STATE Add $ (\mathcal{T'}, y)_{\text{user}_i}$ to $\mathcal{D}'$
        \ENDIF
    \ENDFOR
    \STATE $\mathcal{U}_{\text{orig}} \gets$ Sample $|\mathcal{D}'|$ samples from $\mathcal{U}_{\text{orig}}$
    \FOR{each $\text{user}_i \in \mathcal{U}_{\text{orig}}$}
        \STATE Add $ (\mathcal{T}, y)_{\text{user}_i}$ to $\mathcal{D}'$
    \ENDFOR
    \IF{$m - |\mathcal{D}'| > 0$\COMMENT{cannot evenly split}}
        \FOR{$\text{user}_{i^*}$ in $\mathcal{U}[|\mathcal{D}'|:\ ]$\quad\COMMENT{remaining index}}
        \STATE Add $ (\mathcal{T''}, y)_{\text{user}_{i^*}}$ to $\mathcal{D}'$
        \ENDFOR
    \ENDIF
\ENDFOR
\STATE \textbf{return} $\mathcal{D}'$
\end{algorithmic}
\end{algorithm}

\section{Mitigation Results}\label{app:G-mitigation-result}

Here we replenish the mitigation results in graph-based method in \cref{tab: rcgn_mitigation_results}, confirming the effectiveness of our proposed mitigation strategy.

\begin{table*}[t]
\centering
\begin{tabular}{ll|cc|cc}
\toprule
\multicolumn{2}{c}{\multirow{2}{*}{BotRGCN}} & \multicolumn{2}{c}{{\small Cresci-2015-Data}} & \multicolumn{2}{c}{{\small Twibot-20}} \\
 &  & {\small $\text{Shortcut}_{\text{te}}$} & {\small $\text{Standard}_{\text{te}}$} & {\small $\text{Shortcut}_{\text{te}}$} & {\small $\text{Standard}_{\text{te}}$} \\
\cmidrule(lr){1-6}
\multirow{5}{*}{Sentiments} & $\text{Shortcut}_{\text{tr}}$ & .749 & .869 & .060 & .522 \\
 & AMR+CIGA & .678\;(9\%$\downarrow$) & .838\;(3\%$\downarrow$) & .112\;(87\%$\uparrow$) & .531\;(1\%$\uparrow$) \\
 & $\text{Text-Level}^{*}$ & .885\;(18\%$\uparrow$) & \underline{.935\;(7\%$\uparrow$)} & .277\;(363\%$\uparrow$) & .588\;(12\%$\uparrow$) \\
 & $\text{Dataset-Level}^{*}$ & .877\;(17\%$\uparrow$) & .929\;(6\%$\uparrow$) & .268\;(348\%$\uparrow$) & .594\;(13\%$\uparrow$) \\
 & $\text{Model-Level}^{*}$ & \underline{.890\;(18\%$\uparrow$)} & .929\;(6\%$\uparrow$) & \underline{.512\;(754\%$\uparrow$)} & \underline{.633\;(21\%$\uparrow$)} \\
\cmidrule(lr){1-6}
\multirow{5}{*}{Emotions} & $\text{Shortcut}_{\text{tr}}$ & .903 & .941 & .126 & .539 \\
 & AMR+CIGA & .770\;(14\%$\downarrow$) & .870\;(7\%$\downarrow$) & .232\;(83\%$\uparrow$) & .538\;(0\%$\downarrow$) \\
 & $\text{Text-Level}^{*}$ & .951\;(5\%$\uparrow$) & .969\;(2\%$\uparrow$) & .223\;(76\%$\uparrow$) & .573\;(6\%$\uparrow$) \\
 & $\text{Dataset-Level}^{*}$ & \underline{.956\;(5\%$\uparrow$)} & \underline{.972\;(3\%$\uparrow$)} & .230\;(82\%$\uparrow$) & .558\;(3\%$\uparrow$) \\
 & $\text{Model-Level}^{*}$ & .946\;(4\%$\uparrow$) & .959\;(1\%$\uparrow$) & \underline{.566\;(349\%$\uparrow$)} & \underline{.653\;(21\%$\uparrow$)} \\
\cmidrule(lr){1-6}
\multirow{5}{*}{Topics} & $\text{Shortcut}_{\text{tr}}$ & \underline{.967} & \underline{.968} & .187 & .548 \\
 & AMR+CIGA & .774\;(19\%$\downarrow$) & .850\;(12\%$\downarrow$) & .189\;(0\%$\uparrow$) & .530\;(3\%$\downarrow$) \\
 & $\text{Text-Level}^{*}$ & .938\;(2\%$\downarrow$) & .965\;(0\%$\downarrow$) & .163\;(12\%$\downarrow$) & .562\;(2\%$\uparrow$) \\
 & $\text{Dataset-Level}^{*}$ & .951\;(1\%$\downarrow$) & .958\;(1\%$\downarrow$) & .106\;(43\%$\downarrow$) & .544\;(0\%$\downarrow$) \\
 & $\text{Model-Level}^{*}$ & .900\;(6\%$\downarrow$) & .903\;(6\%$\downarrow$) & \underline{.625\;(234\%$\uparrow$)} & \underline{.704\;(28\%$\uparrow$)} \\
\cmidrule(lr){1-6}
\multirow{5}{*}{Values} & $\text{Shortcut}_{\text{tr}}$ & .815 & .904 & .215 & .564 \\
 & AMR+CIGA & .592\;(27\%$\downarrow$) & .785\;(13\%$\downarrow$) & .225\;(4\%$\uparrow$) & .536\;(4\%$\downarrow$) \\
 & $\text{Text-Level}^{*}$ & .769\;(5\%$\downarrow$) & .881\;(2\%$\downarrow$) & .256\;(19\%$\uparrow$) & .594\;(5\%$\uparrow$) \\
 & $\text{Dataset-Level}^{*}$ & .787\;(3\%$\downarrow$) & .891\;(1\%$\downarrow$) & .246\;(14\%$\uparrow$) & .582\;(3\%$\uparrow$) \\
 & $\text{Model-Level}^{*}$ & \underline{.823\;(0\%$\uparrow$)} & \underline{.905\;(0\%$\uparrow$)} & \underline{.578\;(168\%$\uparrow$)} & \underline{.666\;(18\%$\uparrow$)} \\
\bottomrule
\end{tabular}
\caption{Mitigation effects of our methods on the BotRGCN model. We compare our mitigation strategies against the shortcut setting and representative debiasing method, and report the relative improvement over the original shortcut scenario. The best performance in each group is highlighted with an underline. The results demonstrate that our methods effectively alleviate the impact of shortcut learning, and in most cases, the performance approaches or even matches that of the standard setting.}
\label{tab: rcgn_mitigation_results}
\end{table*}

\section{Fine Tune Discussion}\label{app:H-fine-tune-discussion}

\begin{figure}[t]
\hspace{-0.5cm}
\includegraphics[width=1.1\linewidth]{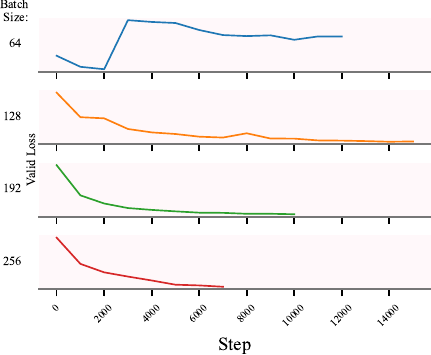}
\caption{Validation loss (y-axis) across training steps for different batch sizes. The loss change rapidly during the initial steps and gradually stabilizes as training progresses.}
\label{fig:val}
\end{figure}

\begin{figure*}[t]
\hspace{-1.75cm}
\includegraphics[width=1.2\linewidth]{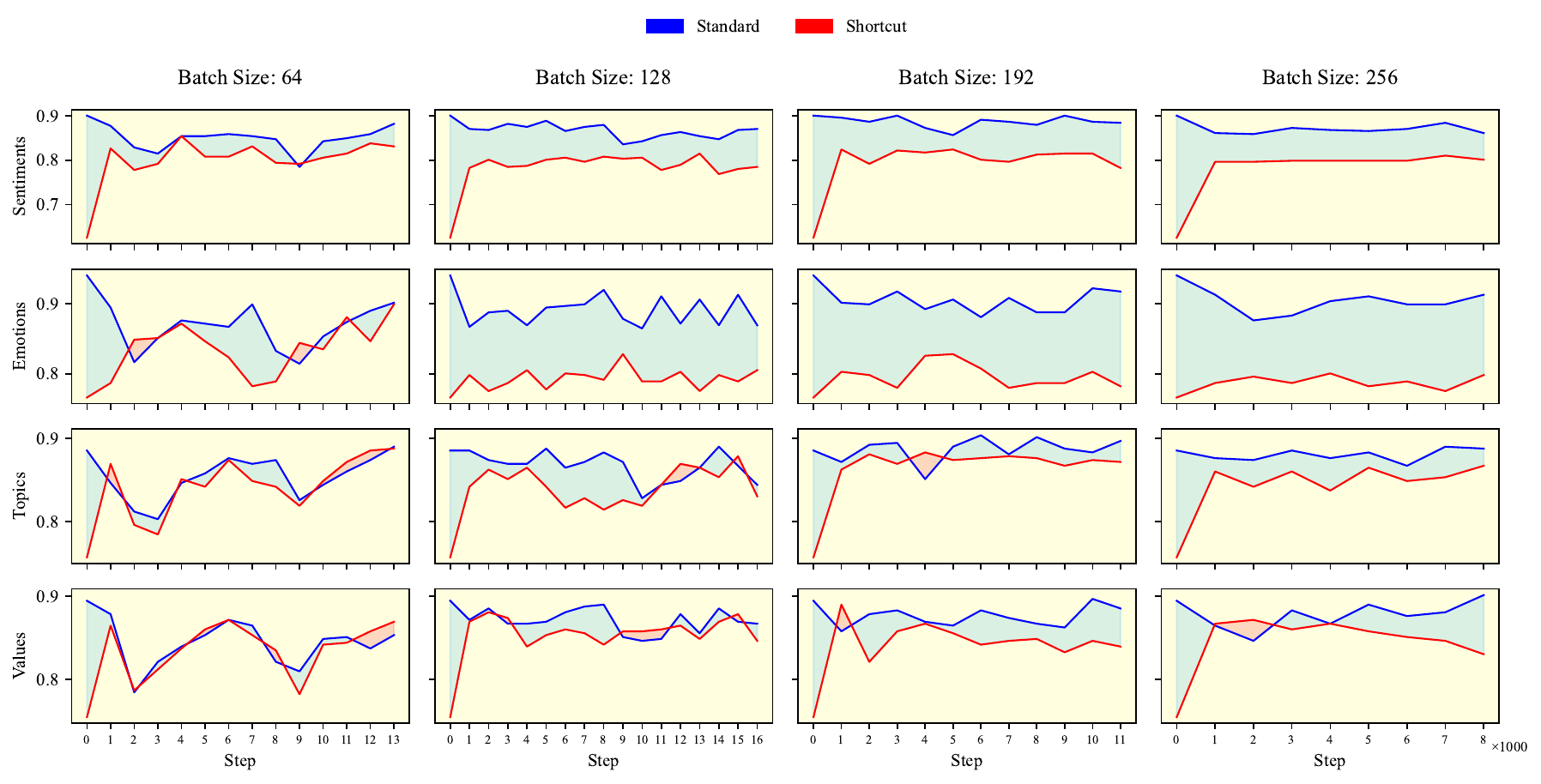}
\caption{Accuracy on the Cresci-2017-Data standard test set using models finetuned with different steps. For each batch size (columns) and feature type (rows), we compare models trained on the standard and shortcut training sets. The leftmost point (step 0) corresponds to the model without finetuning. Results show that finetuning on the standard training set leads to modest changes in performance, while models trained on the shortcut set exhibit a rapid increase in accuracy, indicating their reliance on shortcut features.}
\label{fig:batches}
\end{figure*}

In this section, we investigate how different finetuning parameters, specifically the number of training steps and batch size, affect the performance of the text embedding model in both standard and shortcut scenarios. First, we present the validation loss curve, which shows a rapid change at the beginning followed by a gradual change as training progresses (\cref{fig:val}). Then we evaluate the impact of different training steps across various batch sizes by comparing model performance on standard and shortcut training settings, with evaluation conducted on the standard test set. Results are shown in \cref{fig:batches}, which indicate that models trained on the standard set exhibit relatively stable performance across different steps and batch sizes, with only slight fluctuations. In contrast, models trained on the shortcut set display a rapid increase in test accuracy as finetuning progresses, suggesting that the model quickly overcomes to shortcut features, showing our method's efficiency.

\section{More Discussion}\label{app:I-more-disscuision}

This work explores the robustness of the social bot detection task, inspired by the concepts of shortcut learning and out-of-distribution (OOD) generalization.
The scenarios considered in this study are designed to investigate whether the social bot detection task is affected by shortcut learning. We further discuss the implications of these scenarios in the Discussion section. In real-world applications, the problem is often more challenging, especially as social networks evolve over time, introducing dynamic changes and additional complexity.

\begin{figure}[t]
\hspace{0cm}
\includegraphics[width=1.1\linewidth]{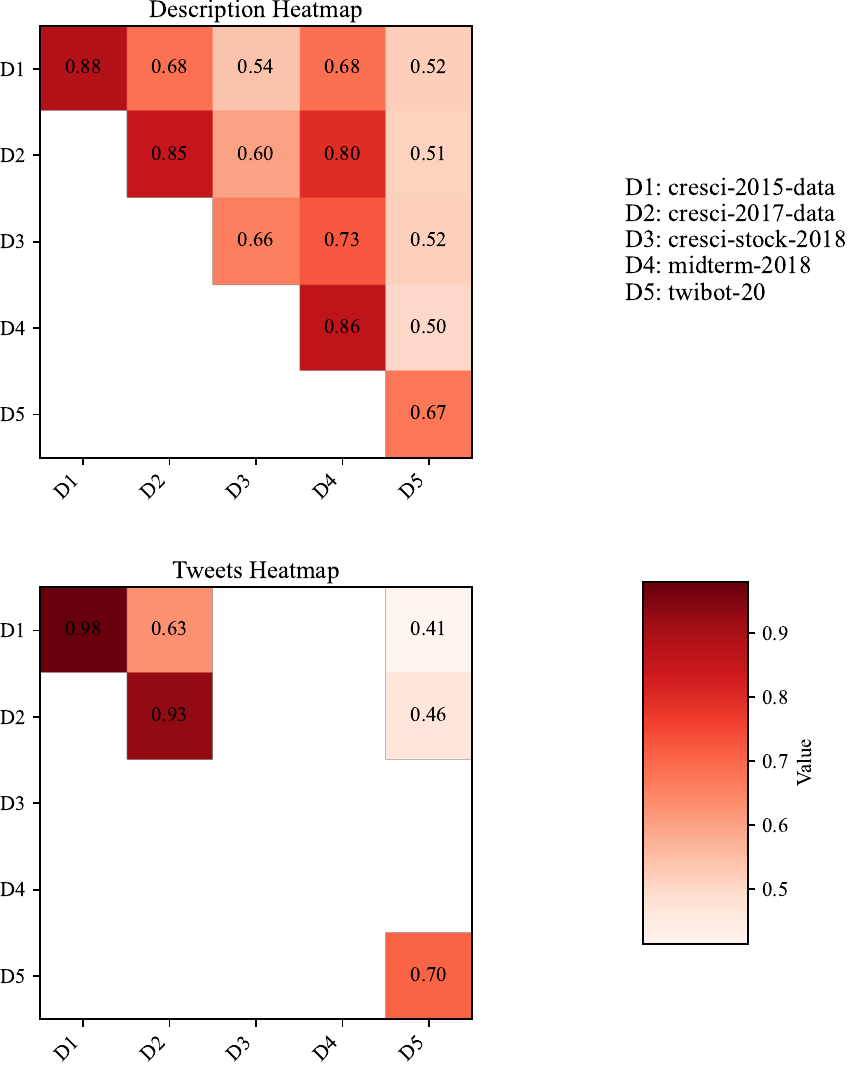}
\caption{Cross-dataset results using the RoBERTa model with an MLP classification head. The upper part reports performance when using descriptions as input features, while the lower part presents results based on tweets (note that datasets D3 and D4 lack tweet information, hence no results are shown for them). The results demonstrate limited generalization ability of the model across datasets.}
\label{fig:heatmap}
\end{figure}

Here, we begin by examining the generalization ability of social bot detectors across datasets. Specifically, we use a RoBERTa model with an MLP classification head and evaluate its performance on several benchmark datasets. As shown in \cref{fig:heatmap}, the model exhibits limited generalization across different datasets.

\begin{figure*}[t]
\hspace{-1.4cm}
\includegraphics[width=1.15\linewidth]{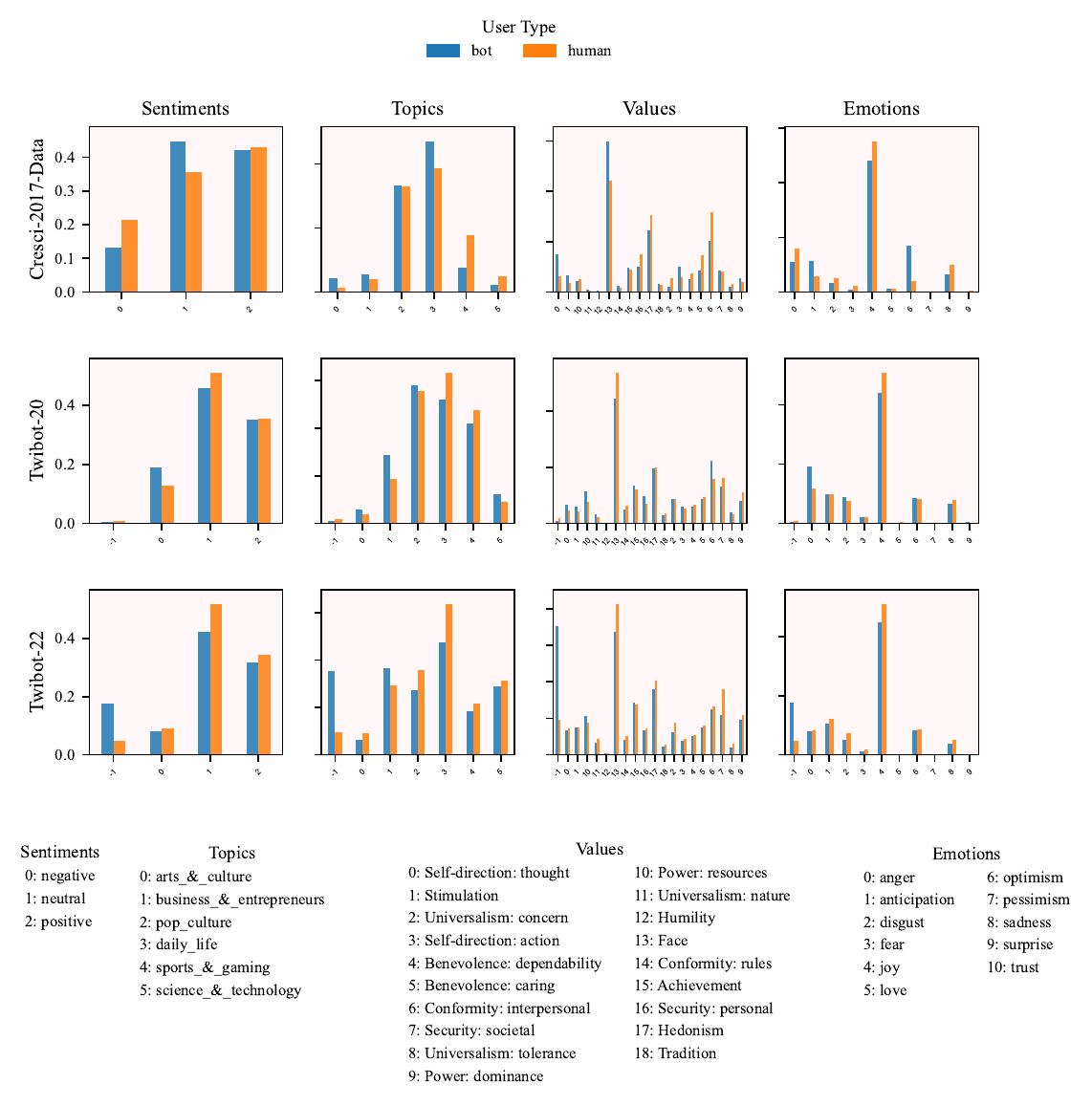}
\caption{Distributional differences of textual features across datasets. The feature category distributions vary significantly across datasets and feature types. In some datasets, social bots and human users exhibit distinct trends under the same feature. Additionally, the same feature demonstrates diverse distributional patterns across different datasets, indicating complex cross-domain differences.}
\label{fig:cross_feature}
\end{figure*}

Additionally, we investigate the distributional differences in shallow textual features across datasets. As illustrated in \cref{fig:cross_feature}, there are complex and significant variations in user-level text characteristics between datasets, further highlighting the challenges of achieving robust and generalizable social bot detection.

\section{Experimental Environment}
\label{app:J-experimental-environment}

All experiments are conducted on Ubuntu 16.04.7 LTS with 377 GB of RAM and NVIDIA GeForce RTX 2080 Ti GPUs.

\end{document}